\documentclass[letterpaper]{article} % DO NOT CHANGE THIS
\usepackage{aaai24}  % DO NOT CHANGE THIS
\usepackage{times}  % DO NOT CHANGE THIS
\usepackage{helvet}  % DO NOT CHANGE THIS
\usepackage{courier}  % DO NOT CHANGE THIS
\usepackage[hyphens]{url}  % DO NOT CHANGE THIS
\usepackage{graphicx} % DO NOT CHANGE THIS
\usepackage{natbib}  % DO NOT CHANGE THIS AND DO NOT ADD ANY OPTIONS TO IT
\usepackage{caption} % DO NOT CHANGE THIS AND DO NOT ADD ANY OPTIONS TO IT
\usepackage{amsmath}
\usepackage{color}
\usepackage{amsfonts}
\usepackage{amssymb}
\usepackage{rotating}
\usepackage{algorithm}
\usepackage{algorithmic}
\usepackage{booktabs}
\usepackage{multirow}
\usepackage{array}
\usepackage{newfloat}
\usepackage{listings}
\usepackage[table]{xcolor}
\DeclareCaptionStyle{ruled}{labelfont=normalfont,labelsep=colon,strut=off} % DO NOT CHANGE THIS
\floatstyle{ruled}
\newfloat{listing}{tb}{lst}{}
\floatname{listing}{Listing}
\title{Towards Robust Image Stitching: An Adaptive Resistance Learning \\against Compatible Attacks}
\author{
	Zhiying Jiang\textsuperscript{\rm 1}, 
	Xingyuan Li\textsuperscript{\rm 1}, 
	Jinyuan Liu\textsuperscript{\rm 2}, 
	Xin Fan\textsuperscript{\rm 1}, 
	Risheng Liu\textsuperscript{\rm 1}\thanks{Corresponding Author.}
}
\affiliations{
	\textsuperscript{\rm 1} Shool of Software Engineering, Dalian University of Technology\\
	\textsuperscript{\rm 2} School of Mechanical Engineering, Dalian University of Technology\\
	\{zyjiang0630, icinesi.li\}@gmail.com, atlantis918@hotmail.com, \{xin.fan, rsliu\}@dlut.edu.cn
}

\usepackage{bibentry}
\begin{document}

\maketitle

\begin{abstract}
	Image stitching seamlessly integrates images captured from varying perspectives into a single wide field-of-view image. Such integration not only broadens the captured scene but also augments holistic perception in computer vision applications. Given a pair of captured images, subtle perturbations and distortions which go unnoticed by the human visual system tend to attack the correspondence matching, impairing the performance of image stitching algorithms. In light of this challenge, this paper presents the first attempt to improve the robustness of image stitching against adversarial attacks. Specifically, we introduce a stitching-oriented attack~(SoA), tailored to amplify the alignment loss within overlapping regions, thereby targeting the feature matching procedure. To establish an attack resistant model, we delve into the robustness of stitching architecture and develop an adaptive adversarial training~(AAT) to balance attack resistance with stitching precision. In this way, we relieve the gap between the routine adversarial training and benign models, ensuring resilience without quality compromise. Comprehensive evaluation across real-world and synthetic datasets validate the deterioration of SoA on stitching performance. Furthermore, AAT emerges as a more robust solution against adversarial perturbations, delivering superior stitching results. Code is available at:~\url{https://github.com/Jzy2017/TRIS}.
\end{abstract}

\section{Introduction}
Image stitching aims to relieve the limitations of camera field-of-view~(FOV) by integrating images from different viewpoints to reconstruct wide FOV scenes. The primary challenge in this task is managing planar transformations in multi-view scenes, especially when aligning a target image with its reference counterpart on a shared plane, achieved through feature matching in overlapping regions.

\begin{figure}[t]
	\centering
	\includegraphics[width=0.45\textwidth]{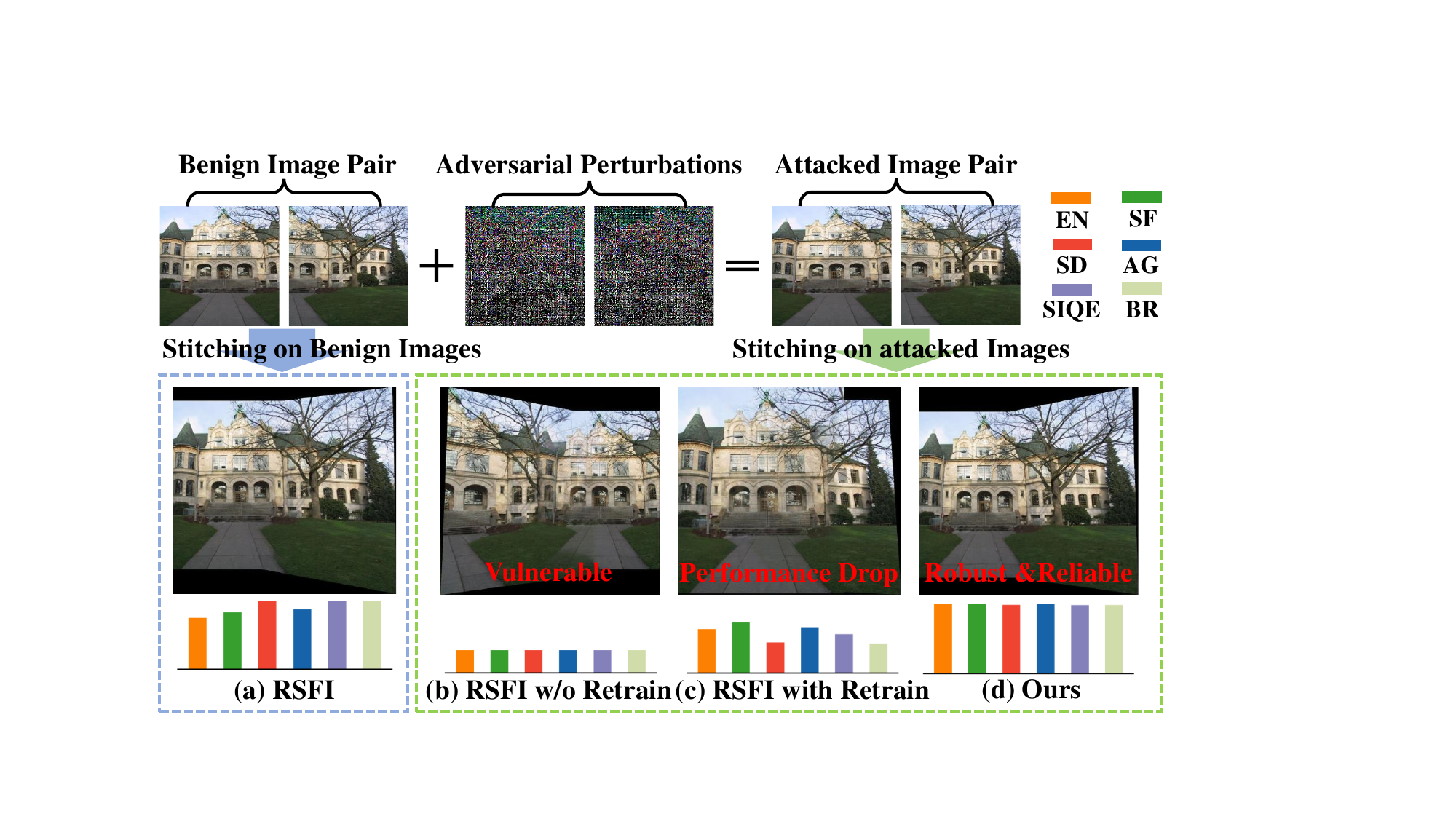}
	\caption{Illustration of our motivation.~(a) presents the reference performance of prolific RSFI~\cite{nie2021unsupervised} on benign images.~(b) reveals the vulnerability of RSFI under adversarial perturbations. Upon routine adversarial training, the robustness of RSFI improves, yet there remains a notable performance decline, as depicted in~(c). In contrast, the proposed method in~(d) not only exhibits resilience against perturbations but also delivers a performance surpassing that observed in the benign scenarios. }
	\label{fig:fig1}
\end{figure}

Early approaches predominantly rely on feature detection, utilizing descriptors such as SIFT~\cite{lowe2004distinctive}, SURF~\cite{bay2006surf}, and ORB~\cite{rublee2011orb}, and established correspondence matching grounded in neighborhood measurements. However, they fall short in cases of the complicated scenarios and often mismatch the detected points. In contrast, the advent of deep learning have ushered in notable advancements in the image stitching~\cite{nie2021unsupervised,jiang2022towards}. These methods harness the rich feature representations offered by deep learning, enabling more precise multi-view alignments. As a result, they yield superior results characterized by minimal ghosting and more trustworthy reconstructions.

While deep learning based image stitching works have achieved significant advancements, their robustness against adversarial attacks remains a concern. Despite their imperceptibility to the human vision, subtle perturbations can drastically modify the predicted results~\cite{liu2023paif}. Given the complexity and diversity of real-world scenes, the imperceptible perturbations can easily blend into the detailed content. However, at the feature level, these perturbations cause marked deviations, severely compromising the accuracy of feature matching in the stitching process. Thus, addressing the vulnerability of image stitching methods to adversarial attacks becomes a critical challenge in ensuring the reliability of the stitched results. 

This paper presents the first endeavor to enhance the robustness of learning based image stitching against adversarial attacks. Specifically, we develop a stitching-oriented attack perturbation~(SoA) tailored for the vital alignment of overlapping content. This perturbation is grounded on an extensive investigation of existing attack strategies, which not only undermines the accuracy of stitching models but also demonstrates compatibility with various prominent attacks. To devise a robust stitching model, we conducted a comprehensive assessment of the adversarial resistance on conventional structures and developed an adaptive architecture search to strike a harmonious equilibrium between attack resistance and stitching accuracy. Through this, the optimal attack resistant architecture is discerned using the adaptive adversarial training~(AAT), mitigating the performance disparities observed in routine adversarial training methods compared to benign models. With the above attack and adaptive training strategies, we construct a robust and flexible stitching framework for challenging and vulnerable applications. Main contributions can be summarized as follows:

\begin{itemize}
	\item This paper advances the robustness of image stitching against adversarial challenges, providing a targeted attack and flexible adversarial training strategy, thereby paving the way for attack resistant image stitching across diverse domains.
	\item Given the deterioration of alignment accuracy, we developed a stitch-oriented attack perturbation compatible with conventional attacks, significantly impairing stitching performance.
	\item We explore the resistance of foundational structure and propose an adaptive adversarial training to determine the robust and effective model, alleviating the performance compromise against the benign models.
	\item Extensive experiments demonstrate that the proposed method achieves a remarkable promotion in both adversarial robustness and stitching performance, outperforming routine adversarial training and benign models.
\end{itemize}

\section{Related Work}
\subsection{Image Stitching}
Existing image stitching methods are mainly developed on feature detection, they employ feature descriptors~\cite{lowe2004distinctive,bay2006surf} to extract feature points and utilize nearest-neighbor, such as RANSAC~\cite{fischler1981random} to match them. By finding the point pairs with similar features in different images, they establish the transformation required for alignment.

Nevertheless, the global consistent transformation often gives rise to the ghosting artifacts.~\cite{gao2011constructing} treated the foreground and background separately to estimate dual homography for the whole images. Smoothly varying affine~\cite{lin2011smoothly} was proposed to address the variable parallax. As-projective-as-possible~\cite{zaragoza2013projective} balanced local nonprojective deviations while adhering to a global projective constraint. Building upon the mesh framework,~\cite{zhang2016multi} enhanced alignment and regularity constraints, enabling support for large baseline and non-planar structures.~\cite{chen2016natural} incorporated a global similarity prior~(GSP) to relieve the local distortion. Furthermore, superpixel based optimal homography was developed~\cite{lee2020warping} to conduct the stitching more adaptively.

More recently, deep learning based image stitching methods have gradually become mainstream.~\cite{nie2020view} introduced a content revision module to address the ghosting and seam artifacts. After that, they proposed ablation constraint to reconstruct the broad scene from feature to pixel~\cite{nie2021unsupervised}.~\cite{song2022weakly} presented a weakly supervised learning method for fisheye panorama generation. Benefiting from the complementary of multi-modality data,~\cite{jiang2022towards,jiang2023multi} proposed infrared and visible image stitching. While the emergence of deep learning has brought about significant improvements in stitching, it also revealed vulnerabilities to adversarial attacks, compromising the accuracy and robustness of the stitched outcomes.
\subsection{Adversarial Attacks}
For deep neural networks, adversarial attacks refer to deceiving the models with perturbed input samples. These perturbations are usually imperceptible to the human vision but compromise the model inference significantly. Recently, various attack methods have been developed. Optimization-based attack method was designed in~\cite{szegedy2013intriguing}, which minimizes the perturbation magnitude while altering the classification results.~\cite{goodfellow2014explaining} proposed a Fast Gradient Sign Method~(FGSM), utilizing model gradients to generate attack examples. However, due to its linear approximation, the examples generated by FGSM are susceptible to detection. To address this,~\cite{kurakin2018adversarial} presented a Basic Iterative Method~(BIM) with multi-step scheme for attack generation.~\cite{madry2017towards} also introduced an iterative scheme, essentially known as Projected Gradient Descent~(PGD). In their approach, gradients of input samples with respect to the loss function are computed, and small perturbations are added in the direction of their gradients.

Some studies on the robustness of the deep learning algorithms have been conducted. In high-level vision tasks,~\cite{Xie_2017_ICCV} attempted to optimize the loss function over a set of targets to improve the adversarial robustness on segmentation and object detection. Based on generative adversarial networks,~\cite{xiao2018generating} enabled the examples share the same distribution with original images, tracing high perceptual quality in defenses.~\cite{Joshi_2019_ICCV} developed an optimization based framework to generate semantically valid adversarial examples using parametric generative transformations to enhance the robustness of classifier. As for low-level tasks,~\cite{yin2018deep} investigated the robust super resolution for different downstream tasks.~\cite{choi2019evaluating} examined the robustness of super resolution methods against adversarial attacks. In image dehazing,~\cite{gao2021advhaze} developed a potentially adversarial haze with high realisticity and misleading.~\cite{yu2022towards} provides a comprehensive analysis on the robustness of existing deraining models. As for image stitching, subtle perturbations decrease the alignment of multi-viewpoints dramatically. However, there is limited works on the vulnerability of deep image stitching methods against adversarial attacks. 
\begin{figure*}[t]
	\centering
	\includegraphics[width=0.98\textwidth]{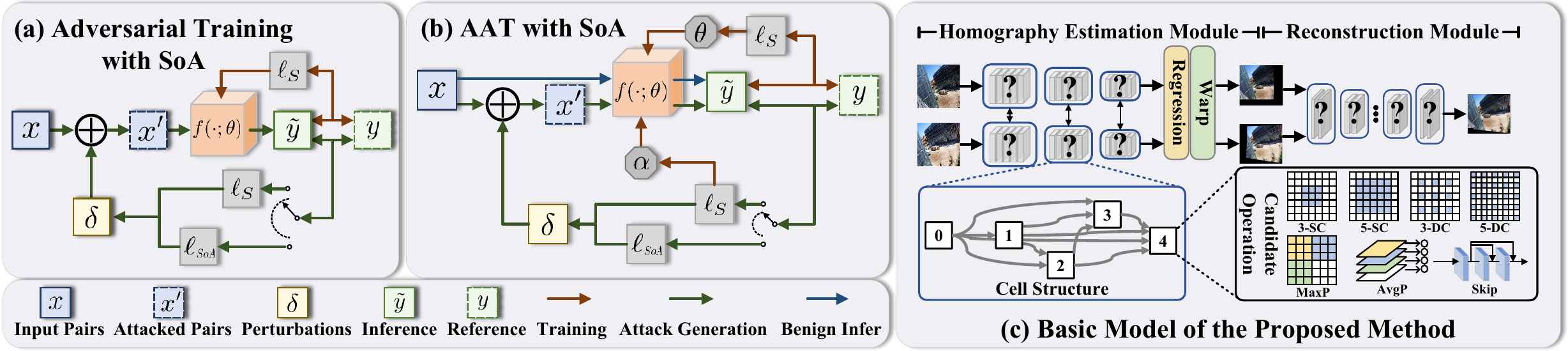}
	\caption{Illustration of the stitching-oriented attacks~(SoA) based routine adversarial training~(a) and the proposed adaptive adversarial training~(AAT)~(b).The basic architecture we employed is shown in~(c).}
	\label{fig:workflow}
\end{figure*}
\section{The Proposed Method}

\subsection{Attacks on Image Stitching}
Adversarial attacks aim to deteriorate the generated wide FOV image by adding subtle, imperceptible perturbations to the input image pairs. To thoroughly challenge network capabilities and to investigate robust and compatible stitching models, we develop a stitch-oriented attack method~(SoA), which focuses on the degradation of alignment across different viewpoints and is instrumental in evaluating the attack resistance of stitching models.

We consider the stitching model~$f(\cdot;\theta)$ parameterized by~$\theta$. Given the input image pair~$(x_1, x_2)$, perturbations~$\delta$ and degradation metric~$M$, the objective of adversarial attacks lies in maximizing the deviation of the generation from the attack-free counterpart, which can be expressed as:
\begin{equation}
	\delta=\mathop{\mathrm{arg\ max}}\limits_{\delta,||\delta||_p\le\epsilon}M(f((x_1,x_2);\theta),f((x_1+\delta,x_2+\delta);\theta)).
\end{equation}
In order to solve this maximization problem under the~$l_p$-bound constraint, we adopt the Projected Gradient Descent~(PGD)~\cite{madry2017towards} and calculate the perturbations in an iterative manner, expressed as:
\begin{align}
	&g = \nabla_{(x_1',x_2')}M(f((x_1,x_2);\theta),f((x_1',x_2');\theta)),\label{eq:equation1}\\
	&(x_1', x_2')=\text{clip}_{[-\epsilon,\epsilon]}((x_1', x_2')+\beta \cdot \textit{sign}(g)),\label{eq:equation2}
\end{align}
where~$(x_1', x_2')\gets(x_1+\delta, x_2+\delta)$,~$\nabla$denotes the gradient operation.
$\beta$ means the step length of each iteration.~$\text{clip}(\cdot)$ guarantees the perturbations are within~$[-\epsilon,\epsilon]$, where~$\epsilon$ represents the maximum perturbation allowed for each pixel.

For image stitching task, the alignment across different viewpoints which primarily relies on image features, is paramount for achieving optimal stitching results. Although the introduced perturbations may remain imperceptible visually, their impact at the feature level is profound. Drawing from this analysis, we develop the corresponding metric~$M$ from the perspective of alignment, incorporating both a supervised loss based on homography and an unsupervised loss grounded in shared region consistency.
\begin{itemize}
	\item \emph{Homography based supervised loss}: homography estimation serves as a conventional procedure for aligning multi-view scenes. We leverage the~${\ell}_2$-norm to quantify the disparity between the homography obtained from the adversarially attacked model and the ground truth, expressed as:
	\begin{equation}
		{\ell}_H = ||H - H'||_{2},
	\end{equation}
	where $H$ is the ground truth homography and~$ H'$ represents the inferred homography from the adversarially attacked model.
	\item \emph{Shared region based unsupervised loss}: the shared content across multi-view scenes provides comprehensive information for assessing alignment accuracy. We only need to focus on the consistency of the shared content between the transformed multiple scenes, which can be formulated as:
	\begin{equation}
		{\ell}_{S} =  ||\mathcal{H}(E) \odot x_1 - \mathcal{H}(x_2)||_{2},   
	\end{equation}
	where $\mathcal{H}(\cdot)$ warps one image to align with the other using estimated homography, $\odot$ is the pixel-wise multiplication and~$E$ is and all-one matrix with identical size with $x_1$. In  practice, for the corresponding measure between the aligned image pairs with/without attacks, we also employ the shared content and the corresponding metric can be represented as:
	\begin{equation}
		{\ell}_{AS} = ||\mathcal{W}(E,I) \odot \mathcal{W}(x_2,H) - \mathcal{W}(E,I) \odot \mathcal{W}(x_2,H')||_{2},  
	\end{equation}
	where~$ I$ and~$H$ are the identity matrix and the estimated homography matrix, respectively. And~$\mathcal{W}(\cdot,\cdot)$ donates the operation of warping an image using a~$3\times3$ transformation matrix with the stitching domain set to the latest large FOV.
\end{itemize}
Accordingly, the corresponding metric~${\ell}_{SoA}$ for the alignment performance between different viewpoints with/without attacks can be defined as: 
\begin{equation}
	{\ell}_{SoA}=  {\ell}_{H} + {\ell}_{AS}.    
\end{equation}
The generation of the proposed stitch-oriented attack~(SoA) perturbations can be summarized in Alg.~\ref{alg:correction1}, where we initialize~$(x_1', x_2') \gets (x_1, x_2)$ and the gradient is calculated on~${\ell}_S$ rather than~${\ell}_{SoA}$ for the first iteration.
\begin{algorithm}
	\caption{SoA based Perturbation Generation}\label{alg:correction1}
	\begin{algorithmic}[1]
		\REQUIRE image pair $(x_1, x_2)$, perturbation bound $\epsilon$,  step size $\beta$, network weights $\theta$
		\STATE $(x_1', x_2') \gets (x_1, x_2)$
		\FOR {$iter$ = 1 to $m$}
		\IF{$iter$ = 1}
		\STATE $g \gets \nabla_{(x_1', x_2')} {\ell}_{S}(\theta,(x_1', x_2'))$
		\ELSE
		\STATE  $g \gets \nabla_{(x_1', x_2')}{\ell}_{SoA}(\theta, (x_1', x_2'), (x_1, x_2))$
		\ENDIF
		\STATE $(x_1', x_2')\gets (x_1', x_2') + \beta \cdot \textit{sign}(g)$
		\STATE $(x_1', x_2') \gets \{(x_1', x_2') \: | \: \|(x_1', x_2') - (x_1, x_2)\|_{\infty} \leq \epsilon\}$
		\ENDFOR
	\end{algorithmic}
\end{algorithm}

\subsection{Adaptive Adversarial Training}
Extensive efforts are currently being devoted to developing specialized learning strategies. Recognizing the intricacies of perturbation generation, many studies have embraced adversarial training to bolster robustness against such attacks. Although integrating attacked data into the training process strengthens resistance, it often compromises task-specific performance~\cite{liu2022twin,jiang2022target}. In order to mitigate the performance degradation exhibited by the models following adversarial training, and to realize an image stitching model with powerful attack-resistance and effective stitching performance, we develop an Adaptive Adversarial Training~(AAT) strategy from architecture perspective.

Specifically, the proposed strategy is developed on differentiable architecture search~(DARTS)~\cite{liu2018darts,liu2022learn,liu2021smoa,liu2023unified}. The differentiable search strategy relaxes the discrete search space into a continuous one by introducing the continuous relaxation~$\alpha$, and the whole optimization objective for search can be formulated as:
\begin{equation}\label{eq:1}
	\begin{split}      
		&\min_\alpha\quad\mathcal{L}_{\rm val}(\alpha;\theta^*)+\lambda\mathcal{L}_{\rm val}^{atk}(\alpha;\theta^*),\\
		&\ {\rm s.t.}\quad\theta^{*}=\arg\min_{\theta}\mathcal{L}_{\rm train}(\theta;\alpha),
	\end{split}
\end{equation}
where~$\mathcal{L}_{\rm train}$,~$\mathcal{L}_{\rm val}$, and~$\mathcal{L}_{\rm val}^{atk}$ denote the training loss, normal validation and attacked validation loss guided with SoA perturbations. The optimization of the aforementioned objective can be decoupled in an iterative manner, focusing separately on the robust architecture training for~$\alpha$ and the standard optimal parameter learning for~$\theta$. Initially, we populate the attacked data with its original counterpart. For the optimization of~$\alpha$, we employ mixed data comprising both normal and attacked samples for standard adversarial training, thereby facilitating robust architecture construction. To balance performance with robustness and prevent search oscillation, we exclusively use the normal data for the weight parameter optimization in the lower objective. The detailed procedure of AAT is delineated in Alg.~\ref{alg3}.

\begin{algorithm}
	\caption{SoA based Adaptive Adversarial Training}\label{alg3}
	\begin{algorithmic}[1]
		\REQUIRE dataset $D$, training epoch $T$, learning rate $\gamma_1$, $\gamma_2$, architecture parameters $\alpha$, network weights $\theta$
		\FOR{epoch = 1, ..., $T$}
		\FOR{minibatch B $\sim$ D}
		\STATE \% Adversarial Examples Generation with SoA
		\FOR{$iter$ = 1 to $m$}
		\STATE Compute project gradient descent $g$
		\STATE $g \gets \mathbb{E}_{(x_1, x_2) \in B}[\nabla_{(x_1', x_2')}({\ell}_{S}(\theta,(x_1',x_2'))\coprod$ \\
		$\qquad \qquad \qquad \quad \:\: {\ell}_{SoA}(\theta,(x_1', x_2'), (x_1, x_2)))]$
		\STATE Update adversarial examples $(x_1', x_2')$ with $g$
		\STATE Project $(x_1', x_2') - (x_1, x_2)$ to $\ell_p$-ball with $\epsilon$
		\ENDFOR
		\STATE \% Architecture Search
		\STATE Compute ${\ell}_{S}(\cdot)$ on $(x_1', x_2')$ with $\theta$
		\STATE $\alpha \gets \alpha - \gamma_1\mathbb{E}_{(x_1, x_2) \in B}[\nabla_\alpha {\ell}_{S} (\theta,(x_1', x_2'))]$
		\STATE \% Weights Learning
		\STATE Compute ${\ell}_{S}(\cdot)$ on $(x_1, x_2)$ with $\alpha$
		\STATE $ \theta \gets \theta - \gamma_2\mathbb{E}_{(x_1, x_2) \in B}[\nabla_\theta {\ell}_{S} (\theta,(x_1, x_2))]$
		\ENDFOR
		\ENDFOR
	\end{algorithmic}
\end{algorithm}

\subsection{Robust Stitching Model}
Our base network is built upon PWCnet~\cite{sun2018pwc}. This network features a three-scale pyramid designed for effective feature encoding and uses an iterative regression mechanism to achieve correspondence matching in a coarse-to-fine manner. In order to enhance the attack-resistance in image stitching, we investigate the robust and flexible structure for feature representation. We predefine a communal cell based on a five-node acyclic graph. Within this graph, each node is associated with a set of mixed operations. Specifically, these operations represent a weighted average of our defined set of candidate operations~\cite{huang2022reconet,liu2020bilevel}. Furthermore, every subsequent node connects with all its predecessors. In this paper, we introduce seven candidate operators, including Skip, Average Pooling~(AvgP), Max Pooling~(MaxP), $3\times3$ SepConv~(3-SC), $5\times5$ SepConv~(5-SC), $3\times3$ Dilated Conv~(3-DC), $5\times5$ Dilated Conv~(5-DC). The employed three-level pyramid is established on three communal cells and the homography can be obtained using the global correlation regression. 

In reconstructing the wide FOV, we employ a U-Net~\cite{ronneberger2015u,liu2023coconet,liu2023holoco} inspired architecture, which incorporates six communal cells. In this setup, the three conventional down-sampling and three up-sampling encoding stages in the traditional U-Net are supplanted by the communal cells. To boost the robustness of our stitching framework, the communal cell is adapted with the similar candidate operator to the homography estimation and the inter connection is determined through the search learning~\cite{li2022learning,li2023gesenet}. A detailed depiction of the structure can be seen in Fig.~\ref{fig:workflow}~(c).

\section{Experiments}
\begin{figure*}[h]
	\centering
	\setlength{\tabcolsep}{1pt}
	\begin{tabular}{cccccccccc}	
		\includegraphics[width=0.06\textwidth,,height=0.09\textheight]{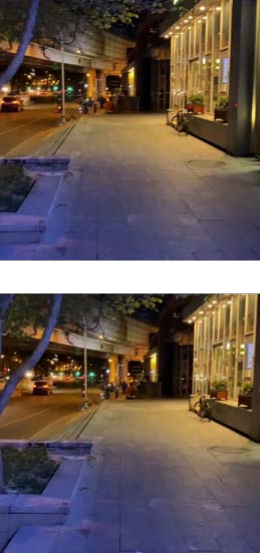}
		&\includegraphics[width=0.17\textwidth,,height=0.09\textheight]{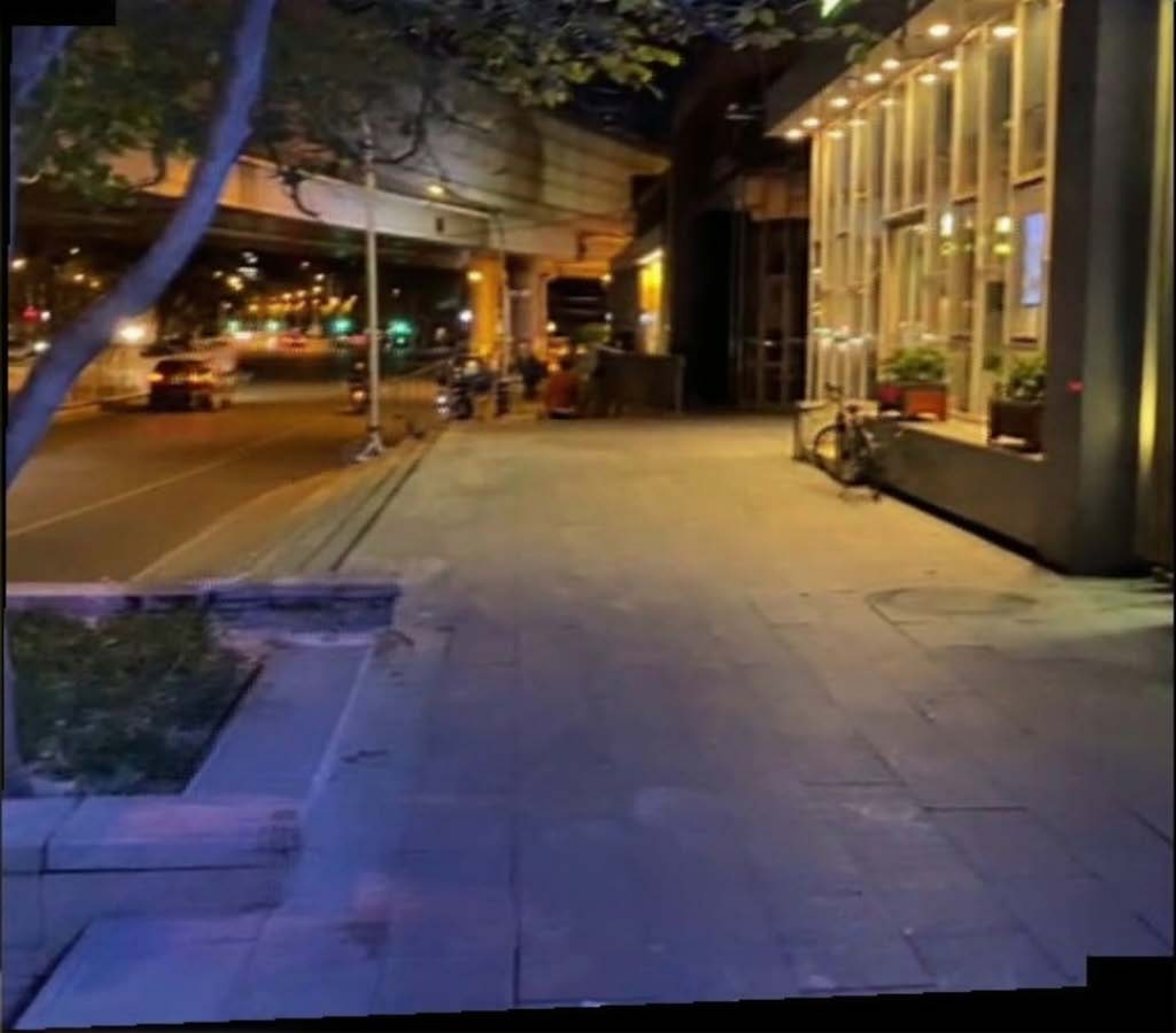}
		&\includegraphics[width=0.17\textwidth,height=0.09\textheight]{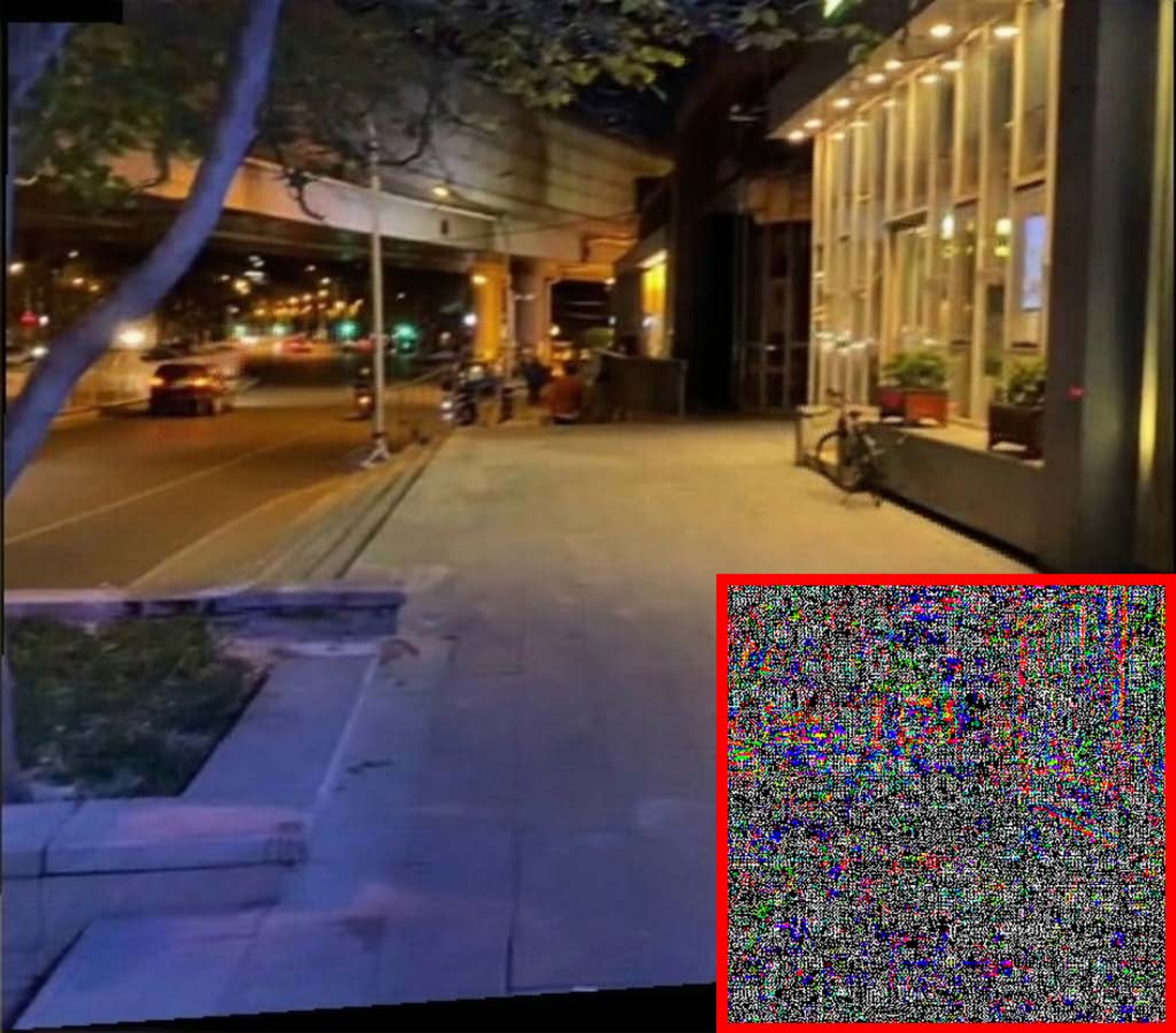}
		&\includegraphics[width=0.17\textwidth,height=0.09\textheight]{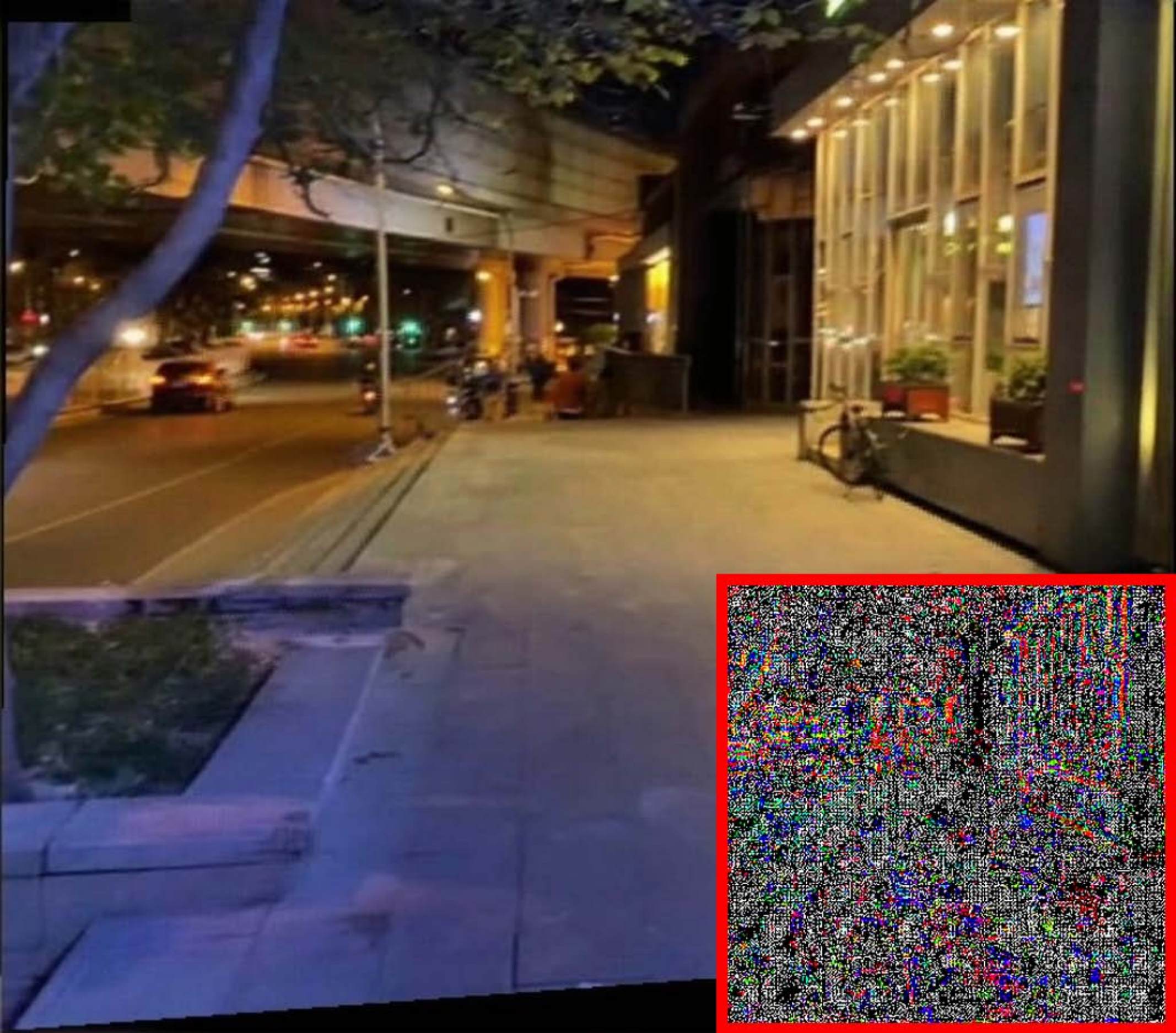}
		&\includegraphics[width=0.17\textwidth,height=0.09\textheight]{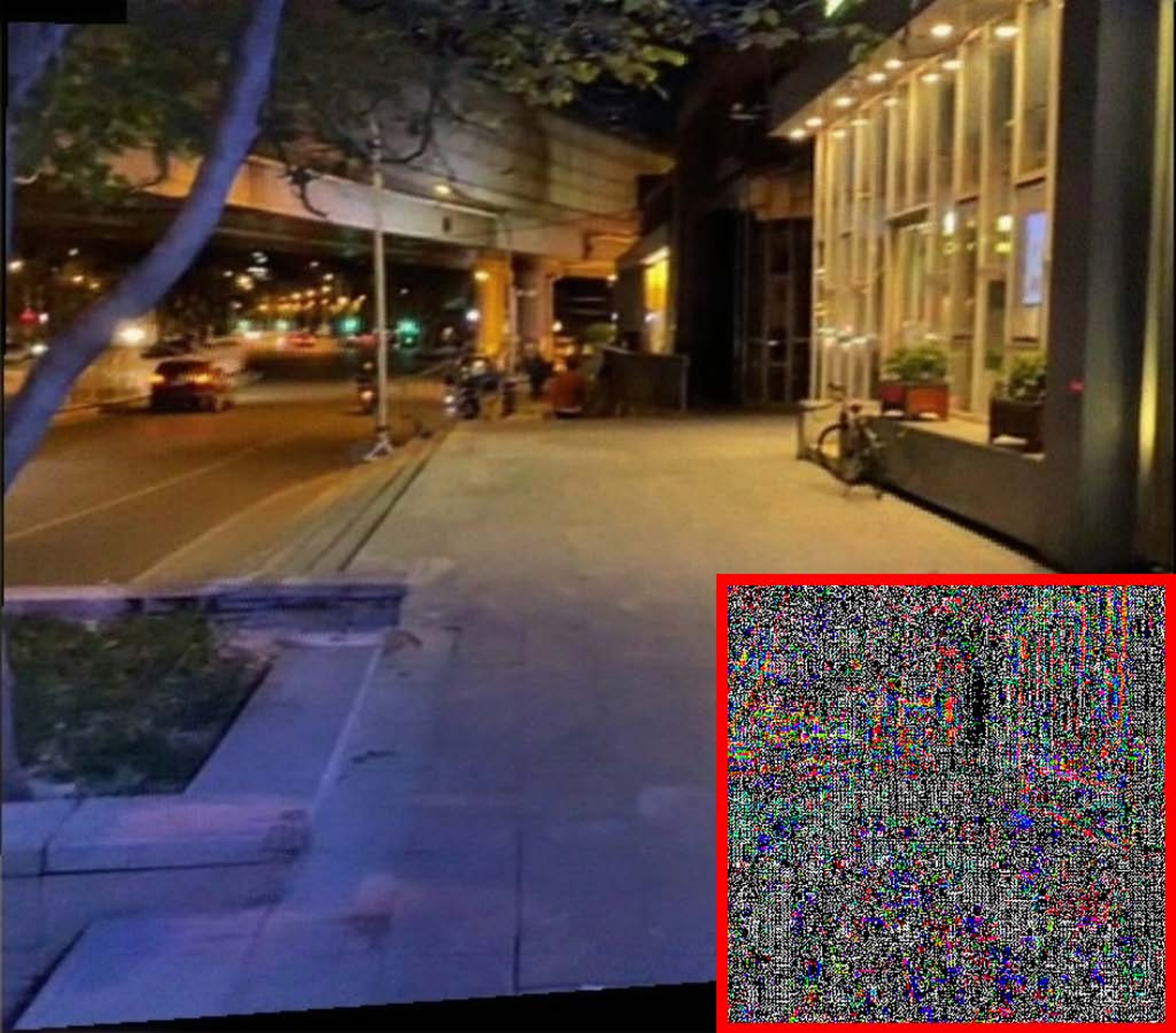}
		&\includegraphics[width=0.17\textwidth,height=0.09\textheight]{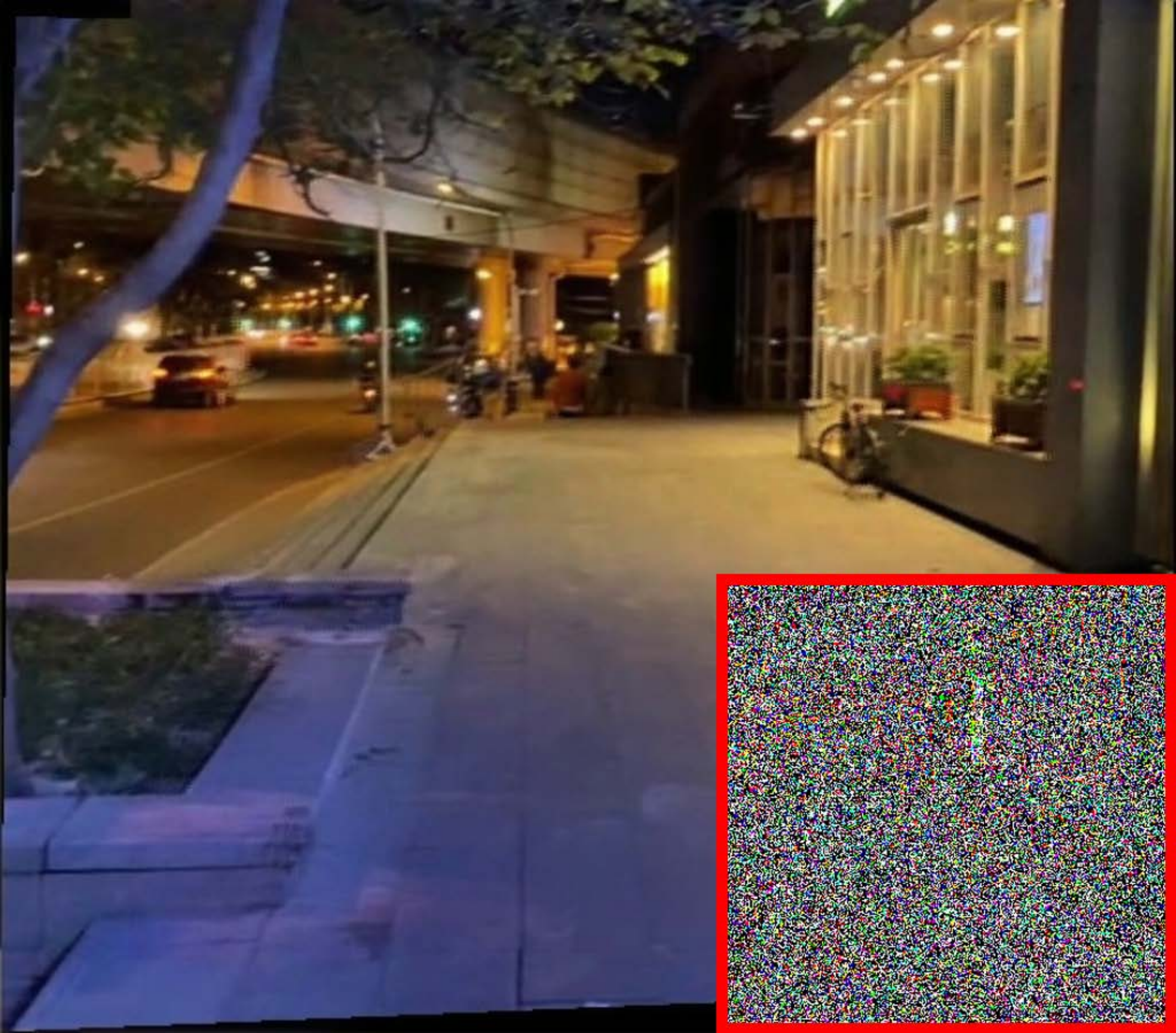}\tabularnewline
		\includegraphics[width=0.06\textwidth,,height=0.09\textheight]{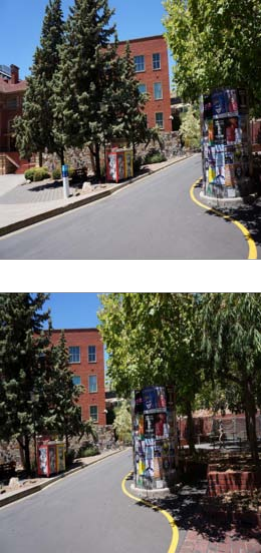}
		&\includegraphics[width=0.17\textwidth,,height=0.09\textheight]{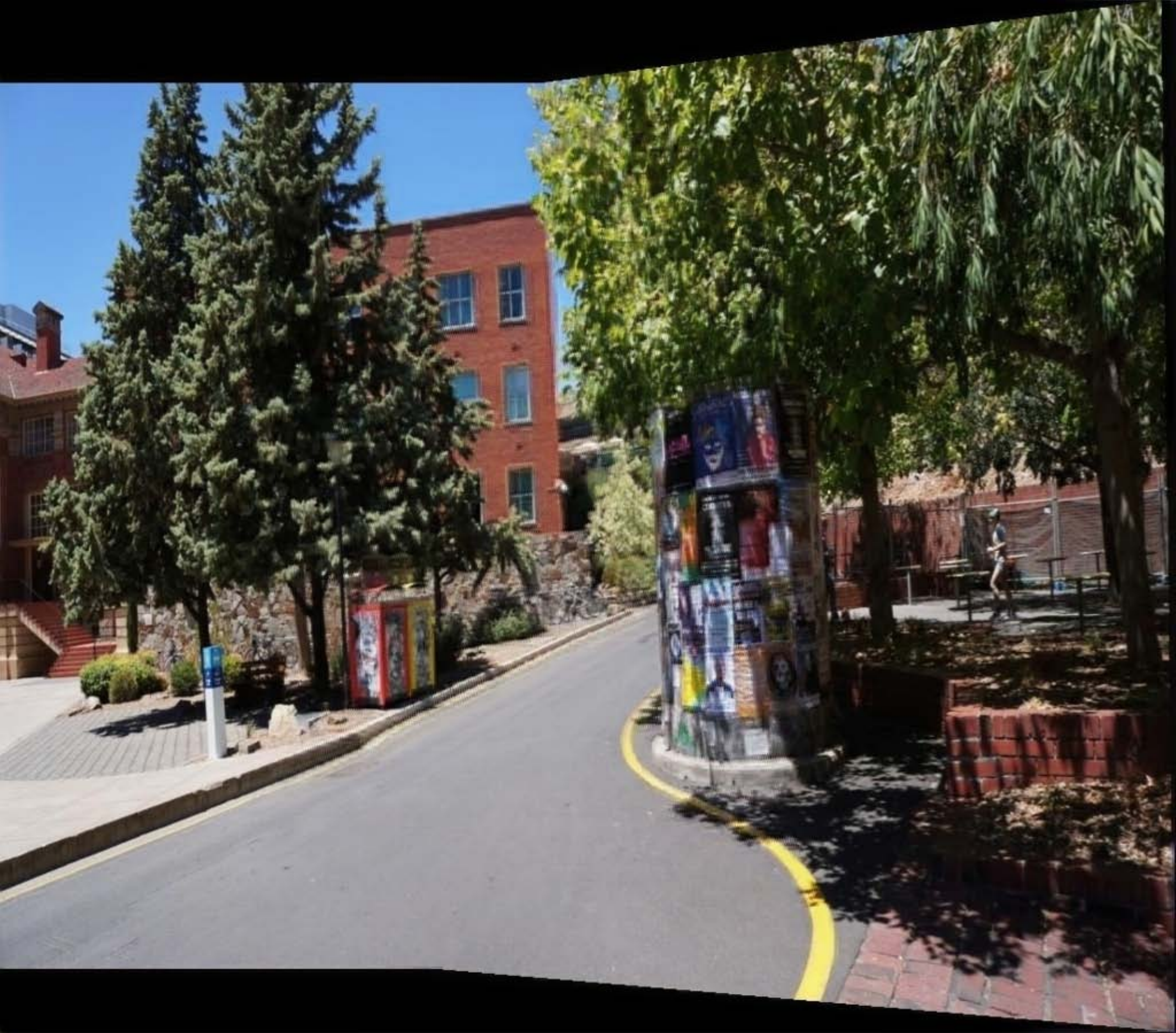}
		&\includegraphics[width=0.17\textwidth,height=0.09\textheight]{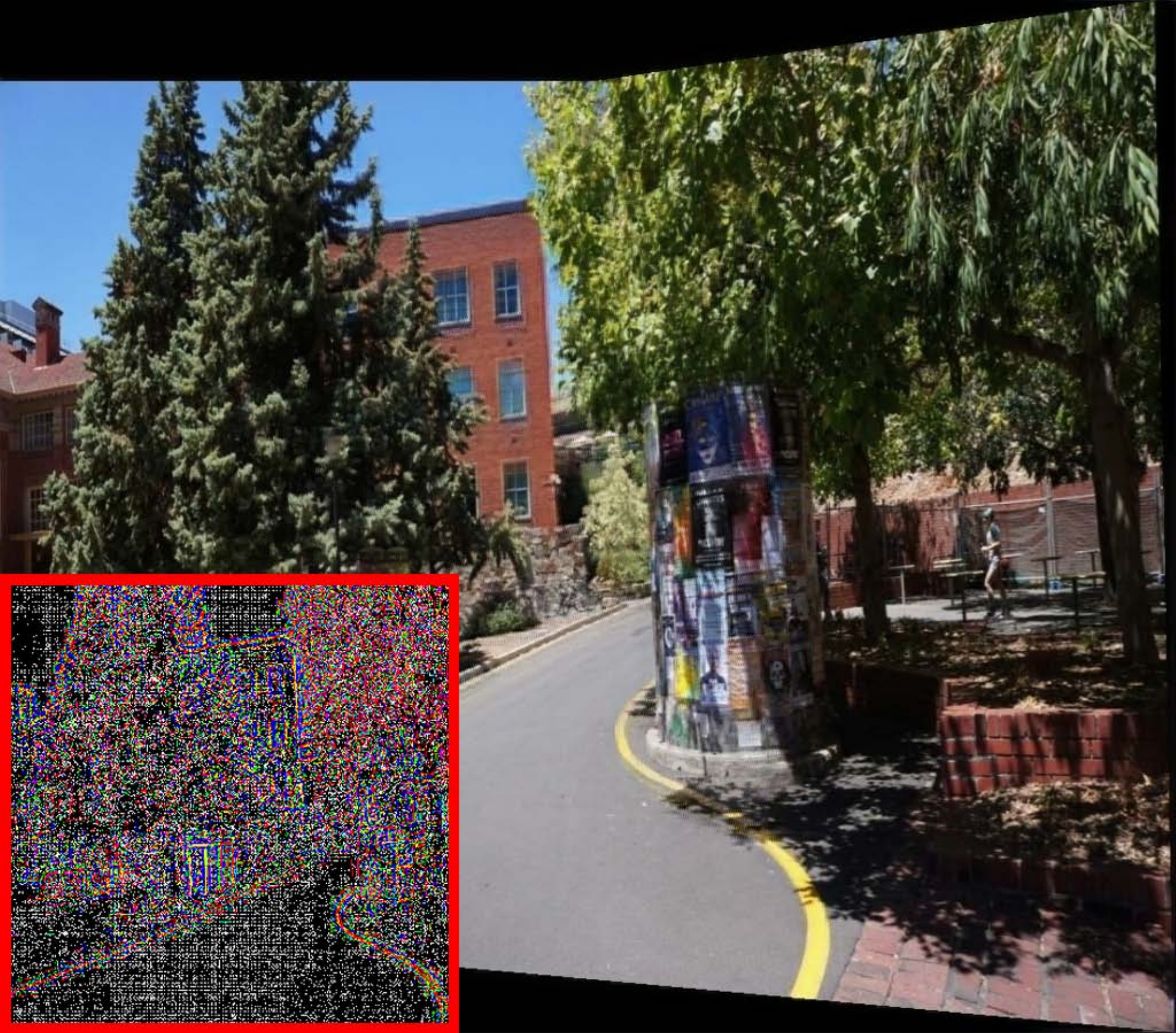}
		&\includegraphics[width=0.17\textwidth,height=0.09\textheight]{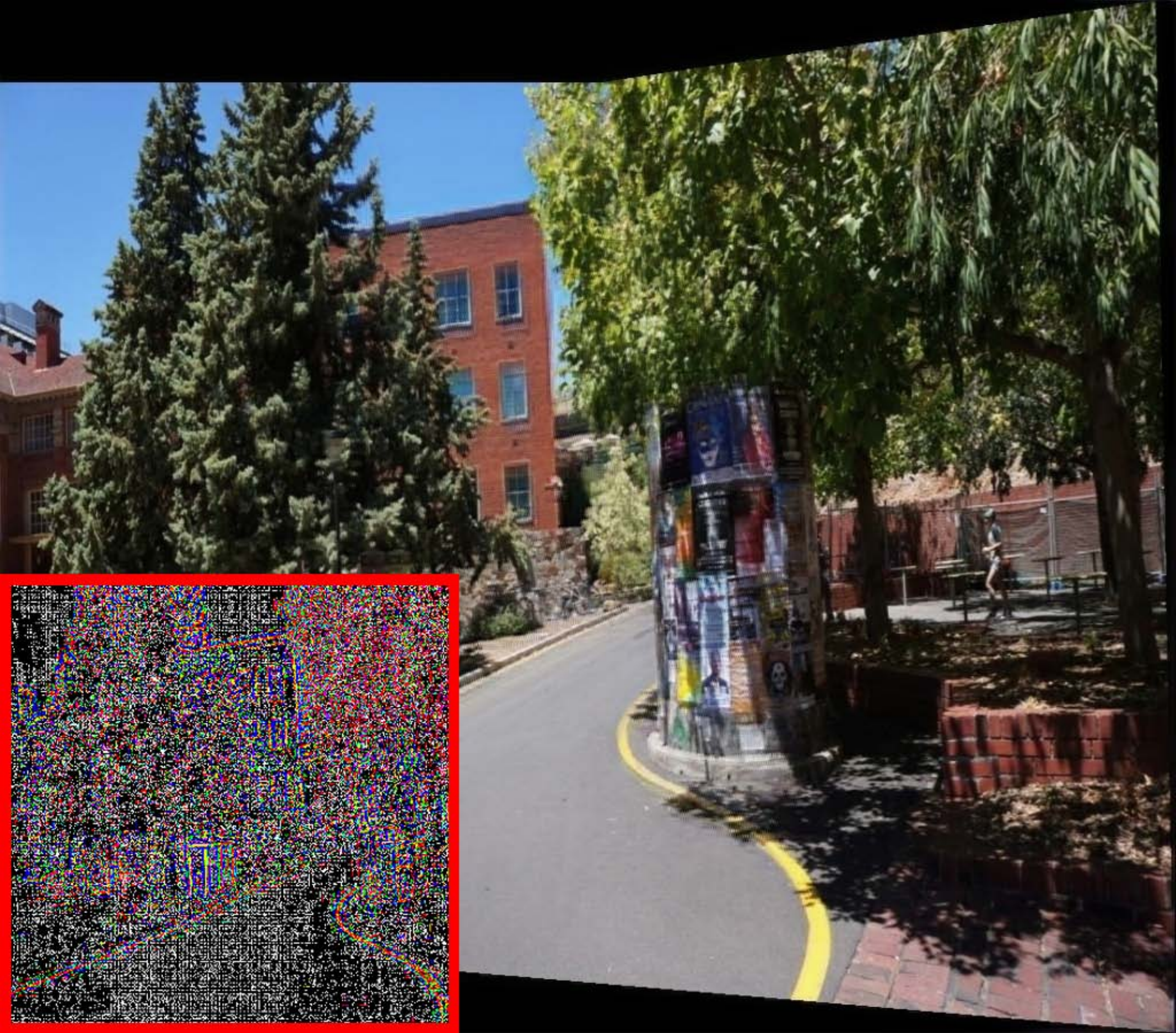}
		&\includegraphics[width=0.17\textwidth,height=0.09\textheight]{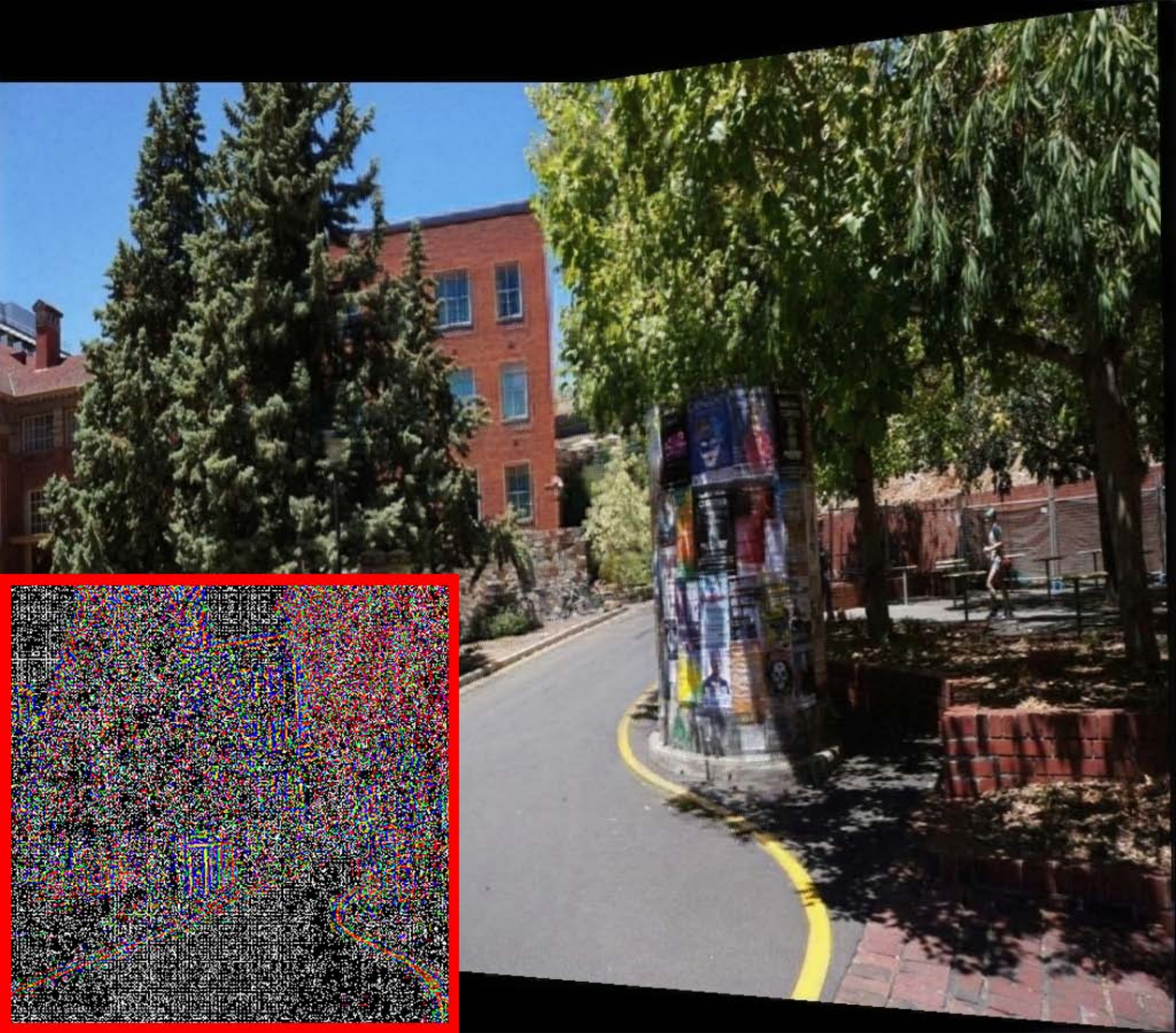}
		&\includegraphics[width=0.17\textwidth,height=0.09\textheight]{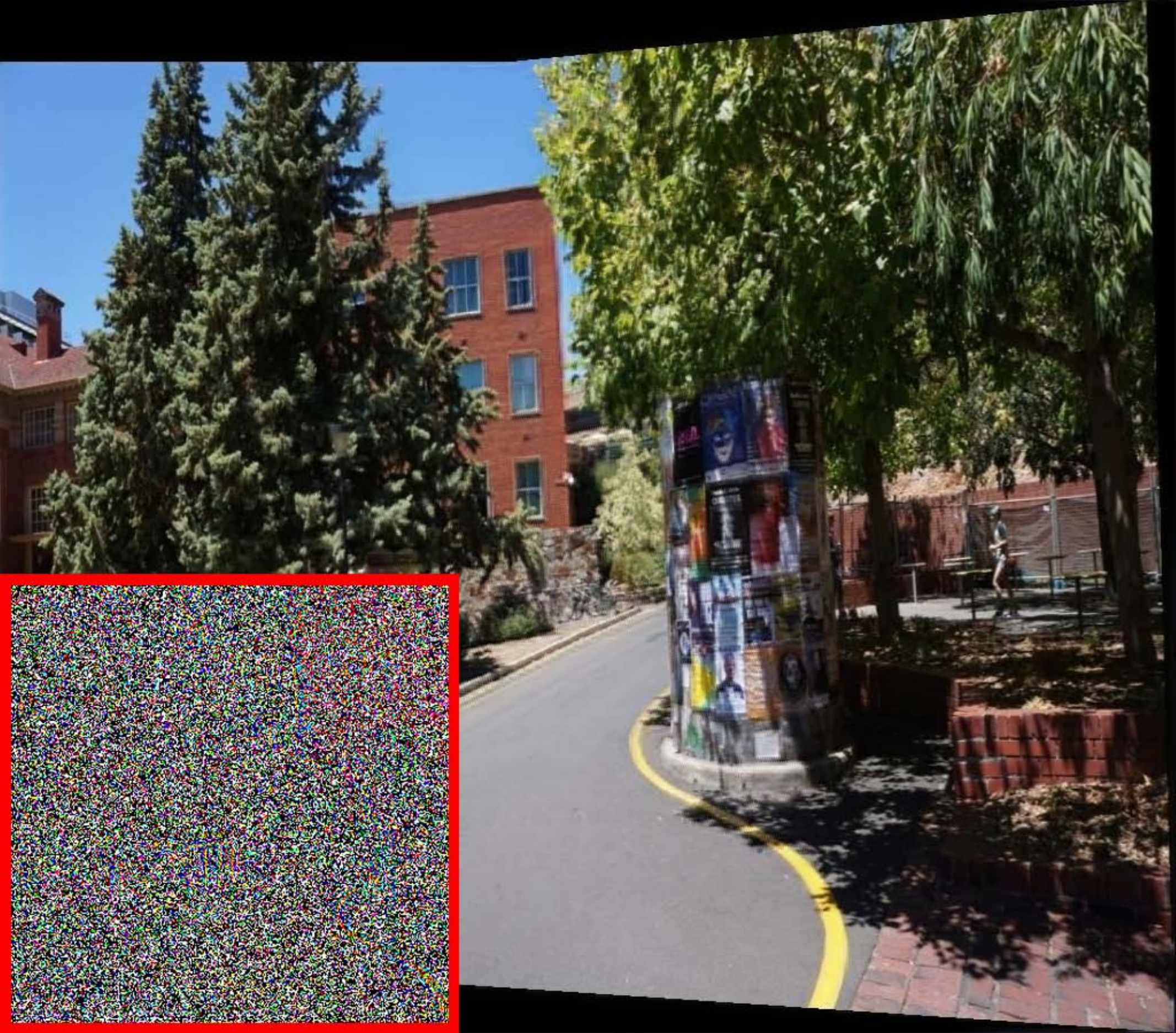}\tabularnewline
		Input&Benign&FGSM&BIM&PGD&SoA\\
	\end{tabular}
%	\vspace{-0.5em}
	\caption{Results of our method under different attacks~(i.e., FGSM, BIM, PGD and SoA ) and in attack-free~(benign) scenarios.}
%	\vspace{-0.5em}
	\label{fig:exp7}
\end{figure*}
\begin{figure*}[h]
	\centering
	\setlength{\tabcolsep}{1pt}
	\begin{tabular}{ccccccccccc}	
		\rotatebox{90}{\ Benign}&\includegraphics[width=0.05\textwidth,height=0.06\textheight]{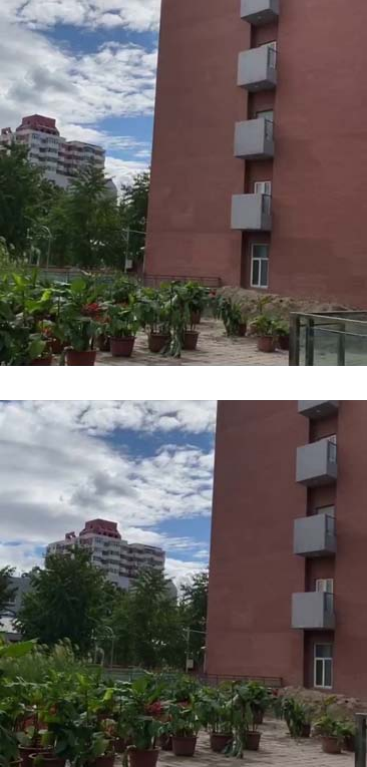}
		&\includegraphics[width=0.13\textwidth,height=0.06\textheight]{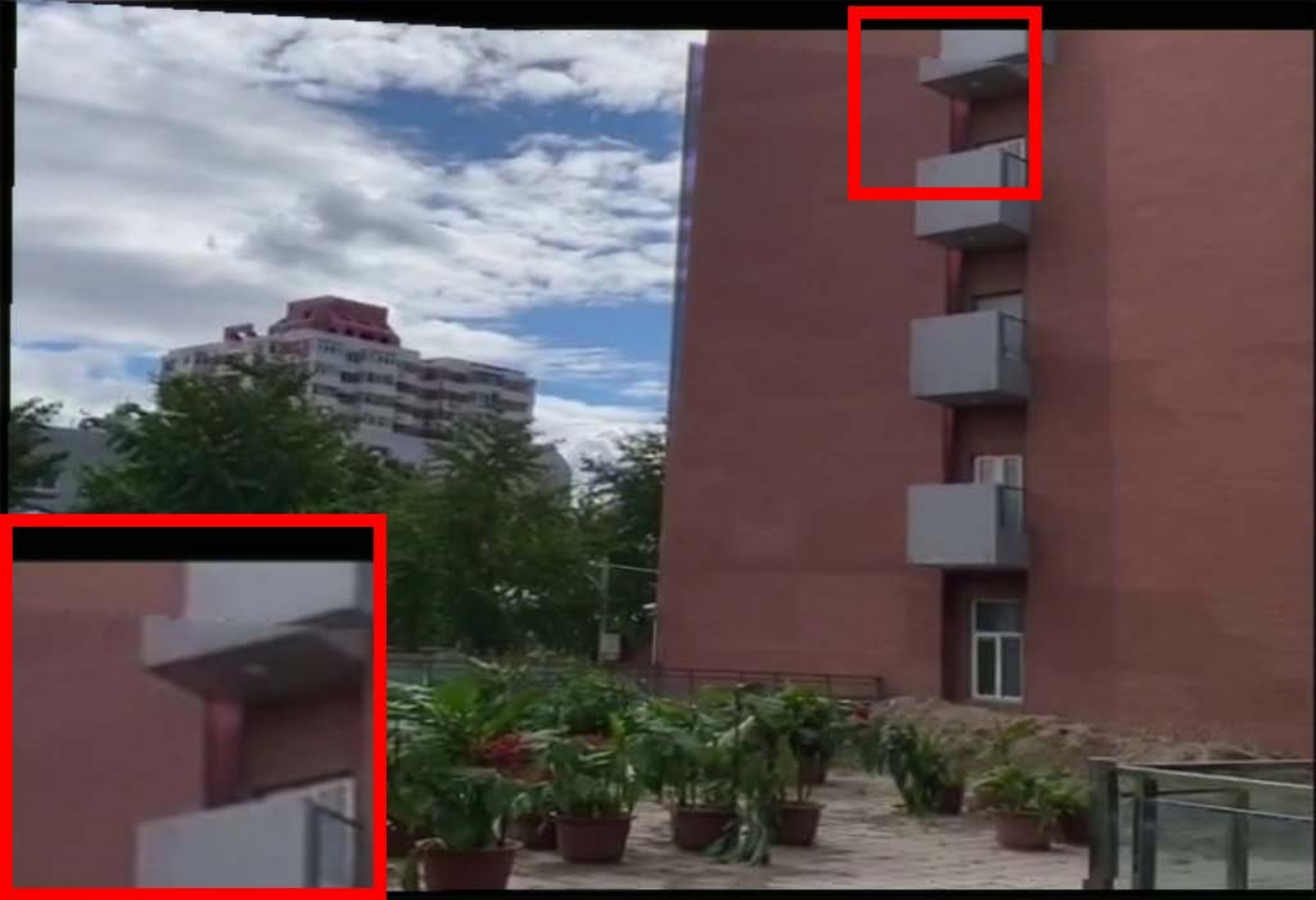}
		&\includegraphics[width=0.13\textwidth,height=0.06\textheight]{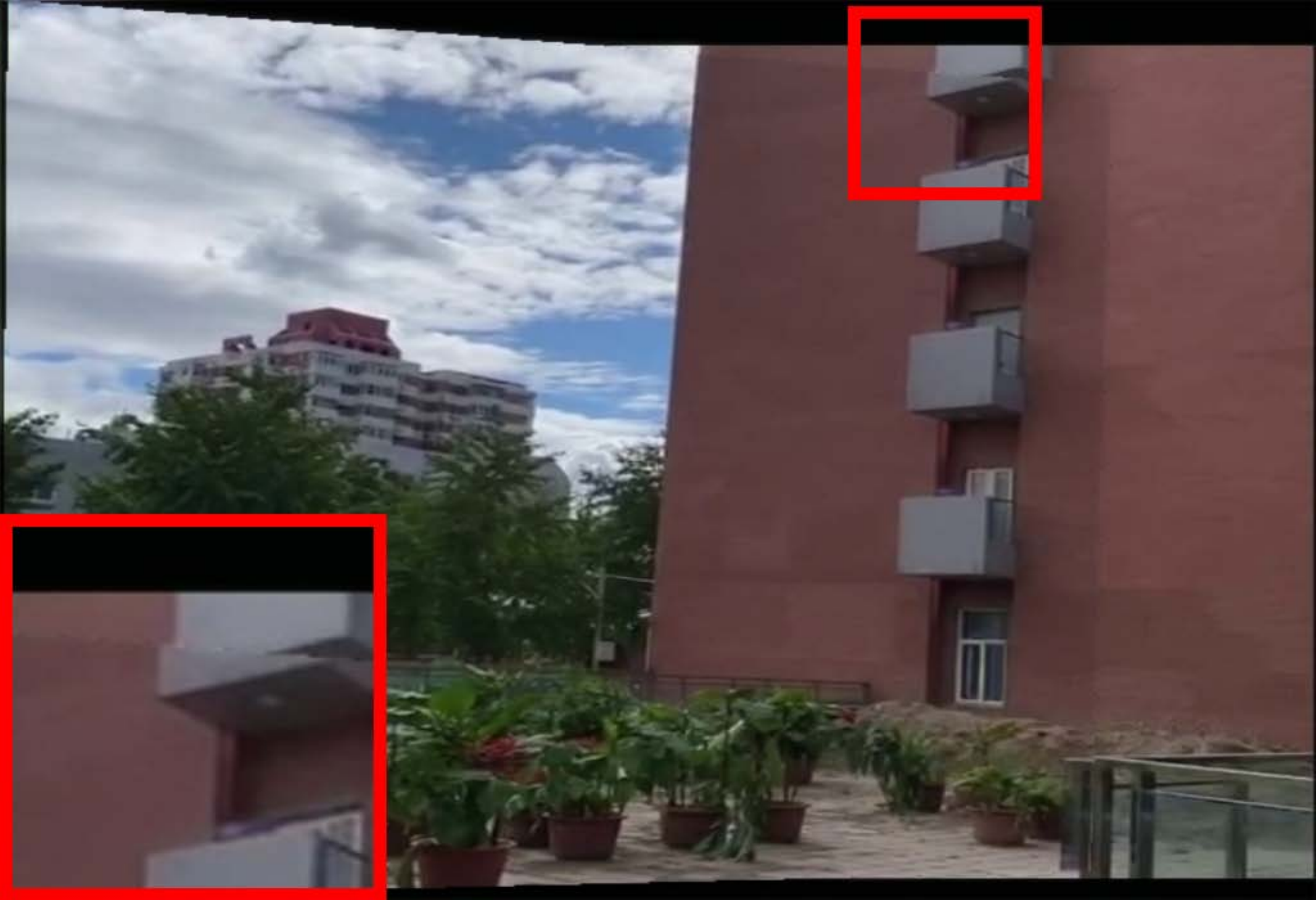}
		&\includegraphics[width=0.13\textwidth,height=0.06\textheight]{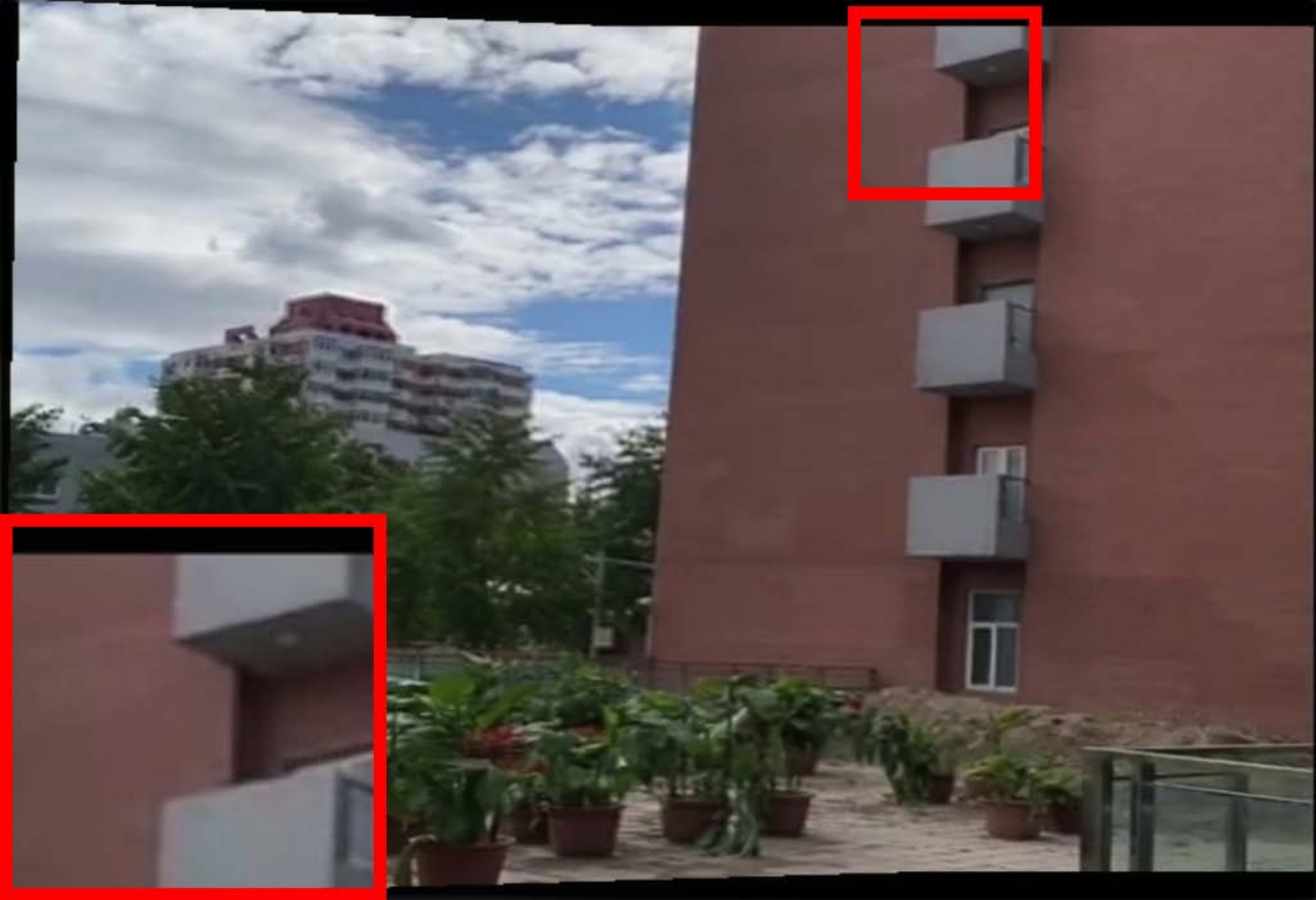}&&
		&\includegraphics[width=0.05\textwidth,height=0.06\textheight]{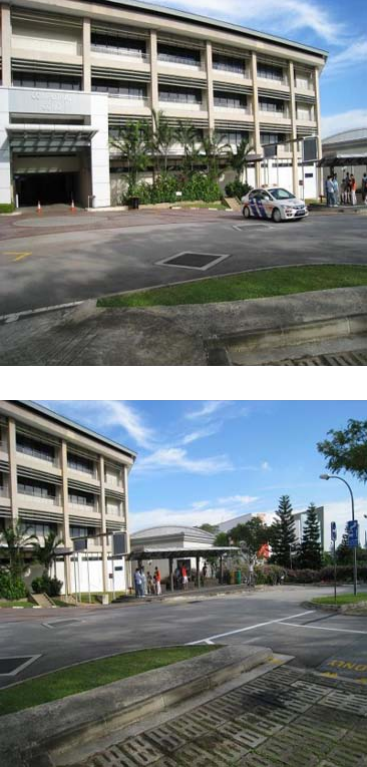}
		&\includegraphics[width=0.13\textwidth,height=0.06\textheight]{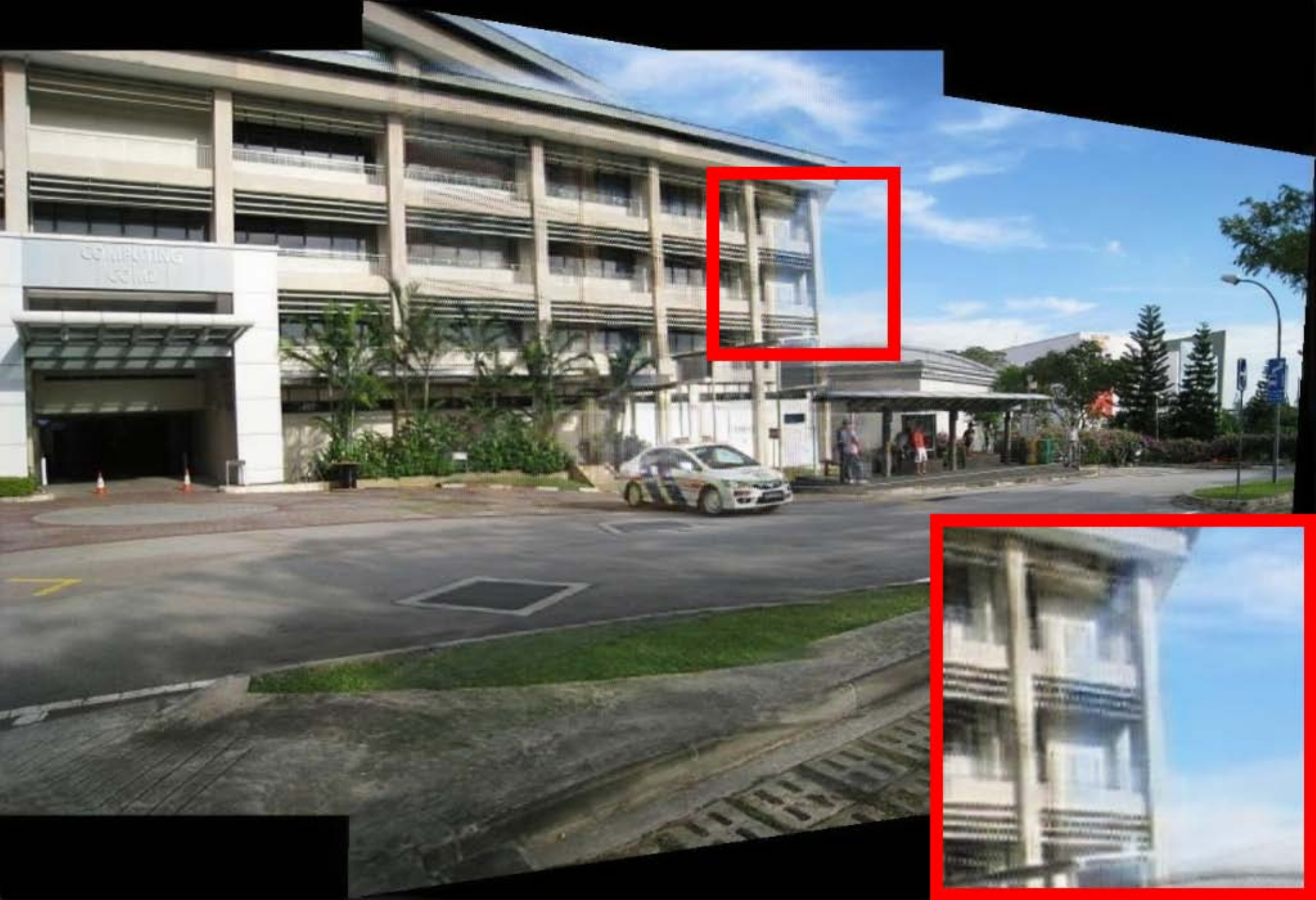}
		&\includegraphics[width=0.13\textwidth,height=0.06\textheight]{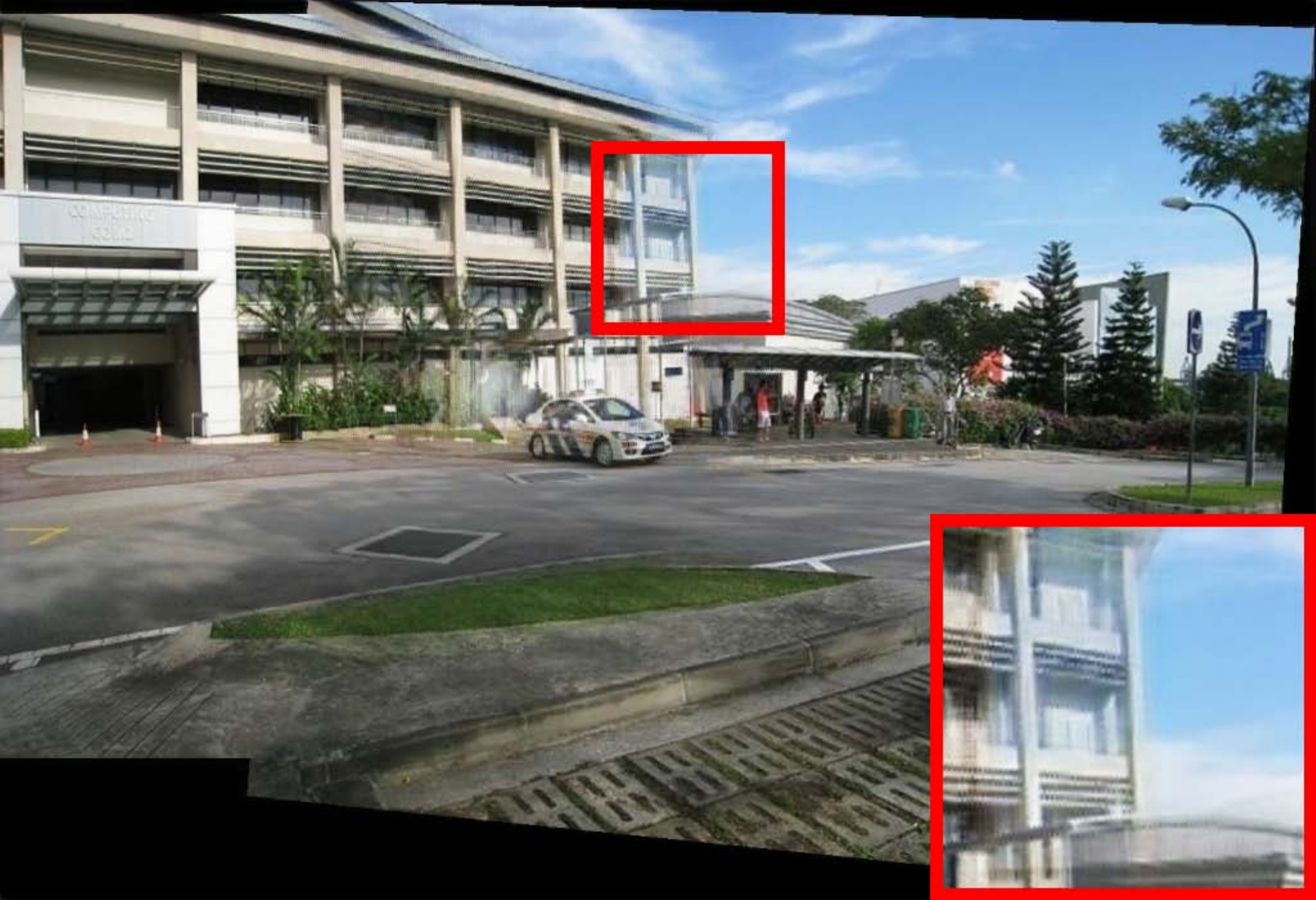}
		&\includegraphics[width=0.13\textwidth,height=0.06\textheight]{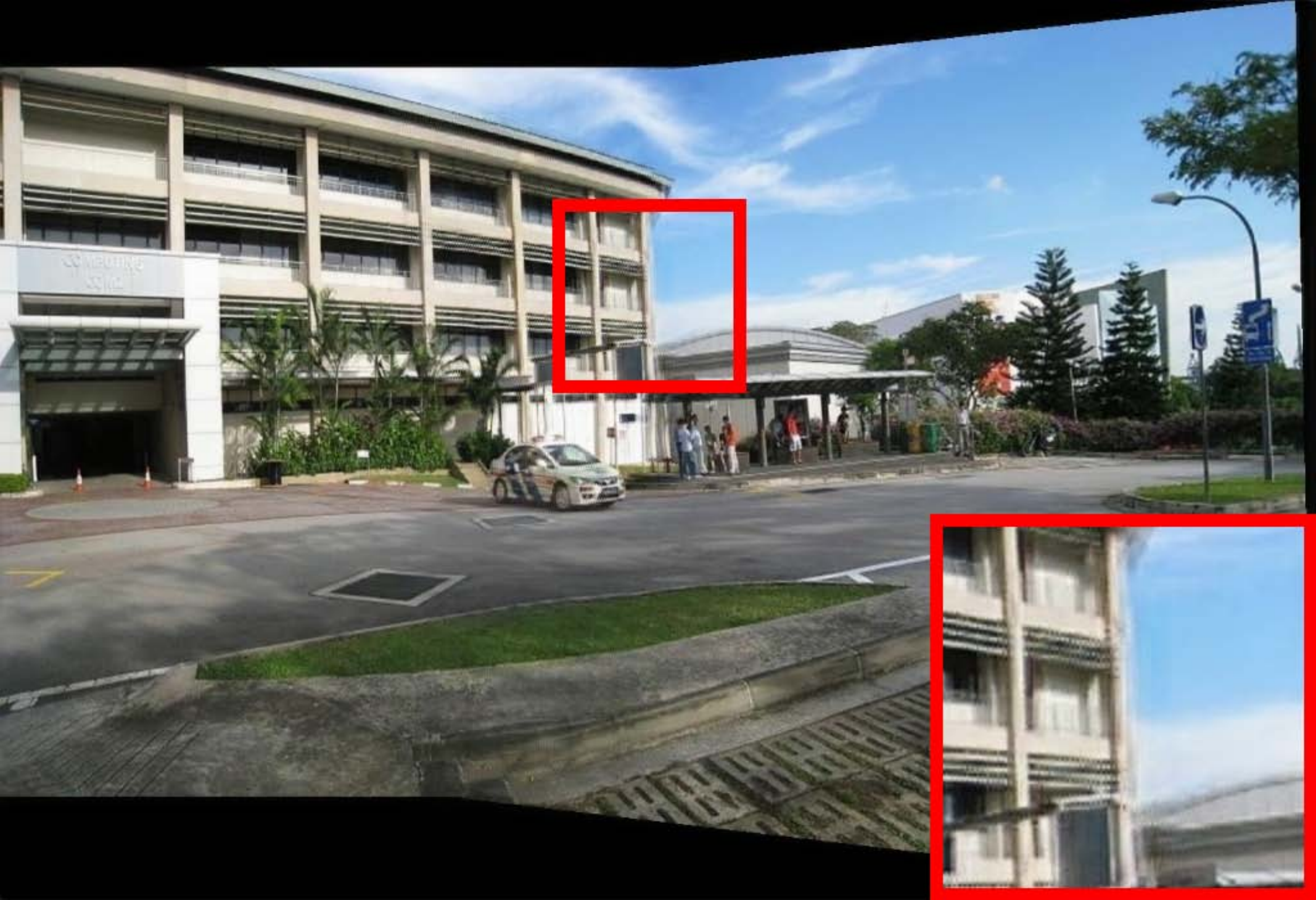}
		\tabularnewline
		\rotatebox{90}{ Attacked}&\includegraphics[width=0.05\textwidth,height=0.06\textheight]{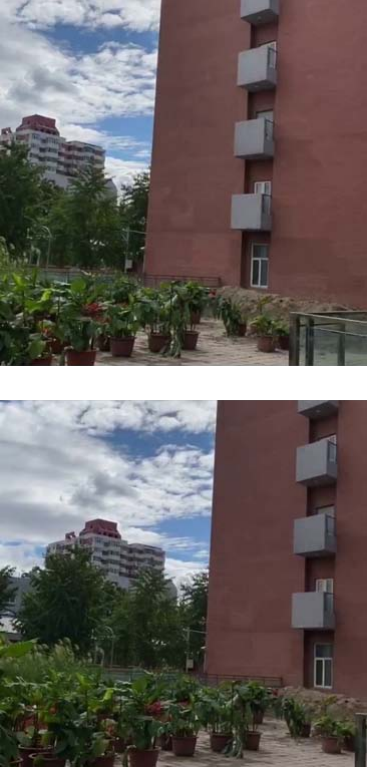}
		&\includegraphics[width=0.13\textwidth,height=0.06\textheight]{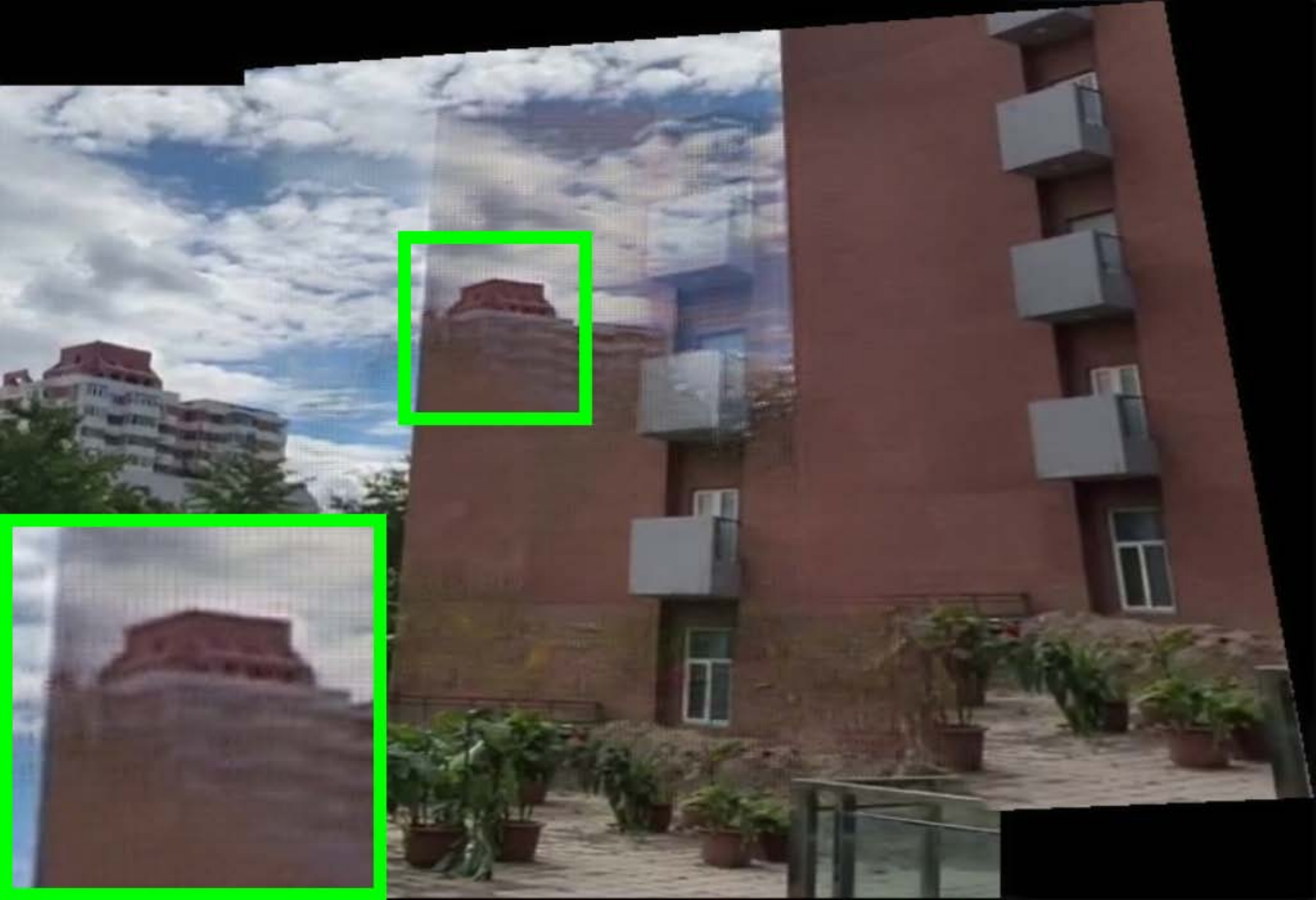}
		&\includegraphics[width=0.13\textwidth,height=0.06\textheight]{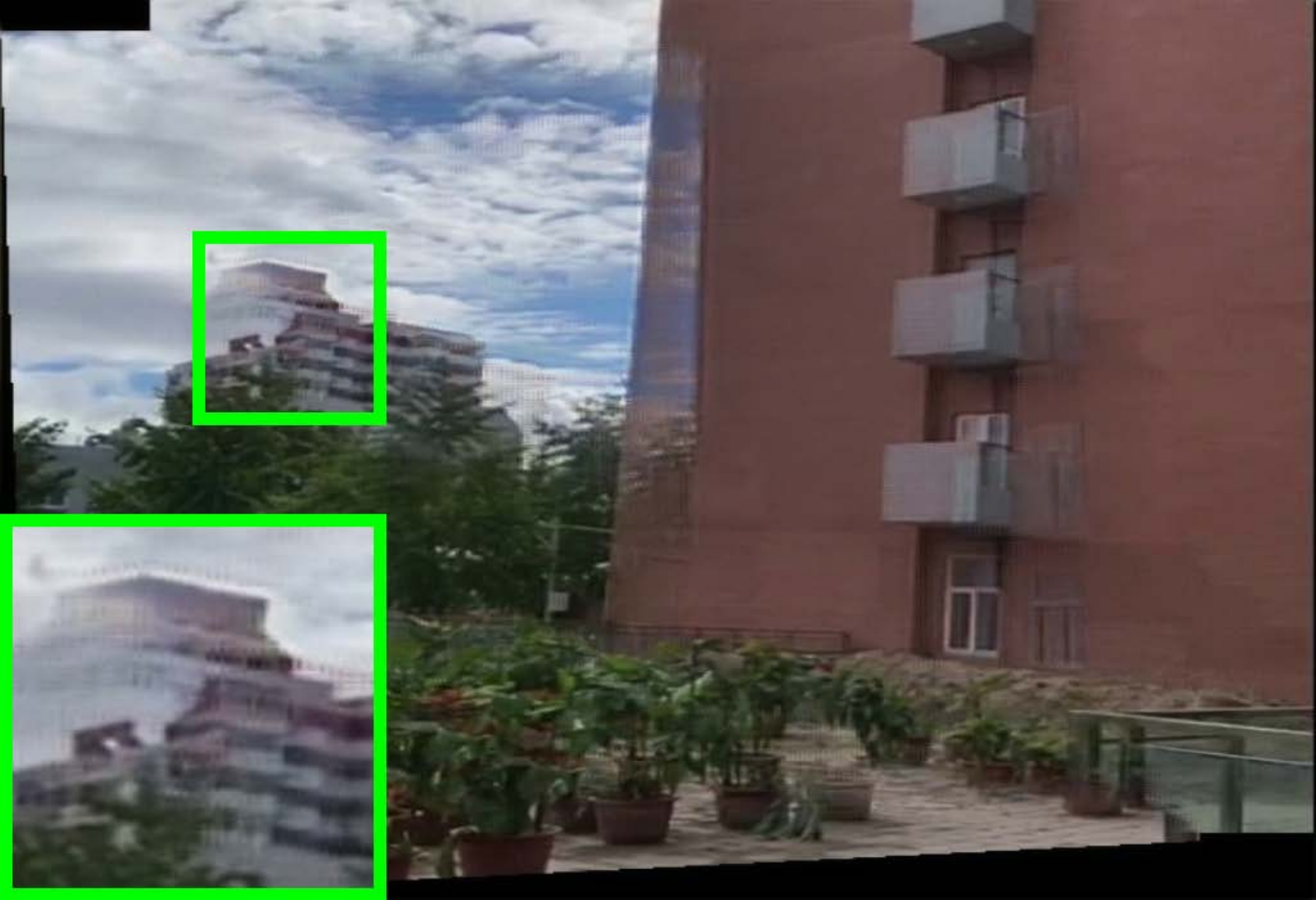}
		&\includegraphics[width=0.13\textwidth,height=0.06\textheight]{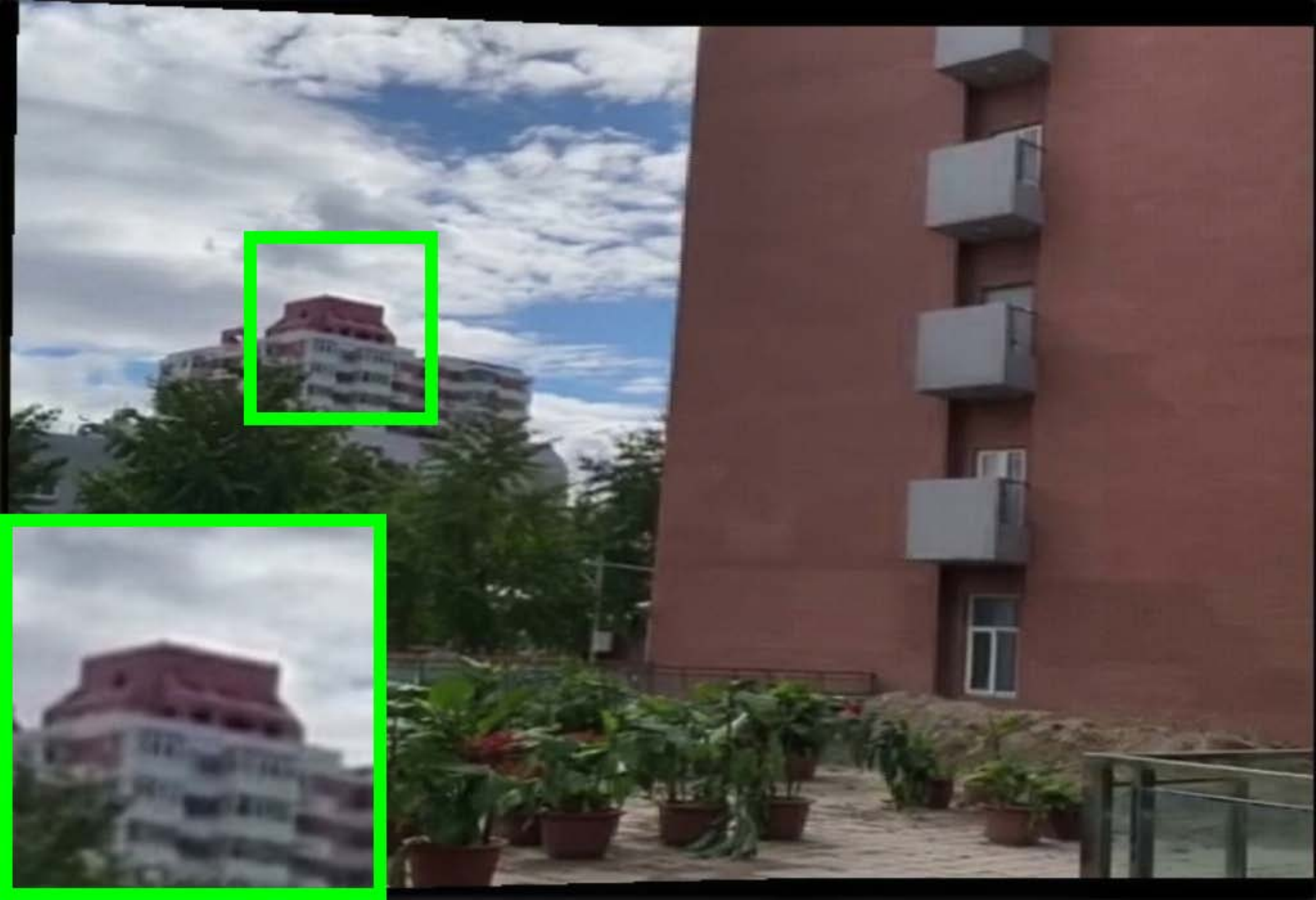}&&
		&\includegraphics[width=0.05\textwidth,height=0.06\textheight]{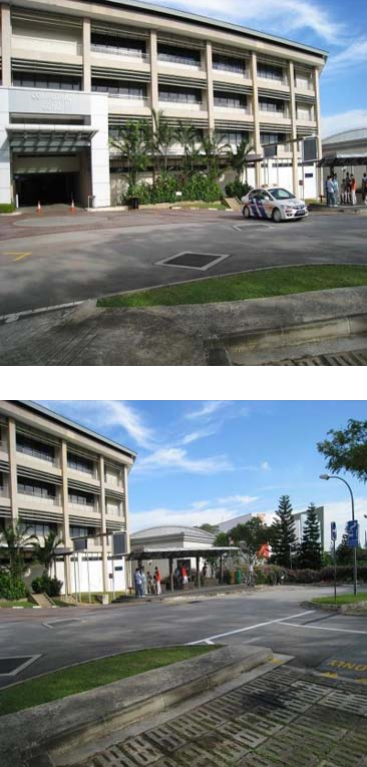}
		&\includegraphics[width=0.13\textwidth,height=0.06\textheight]{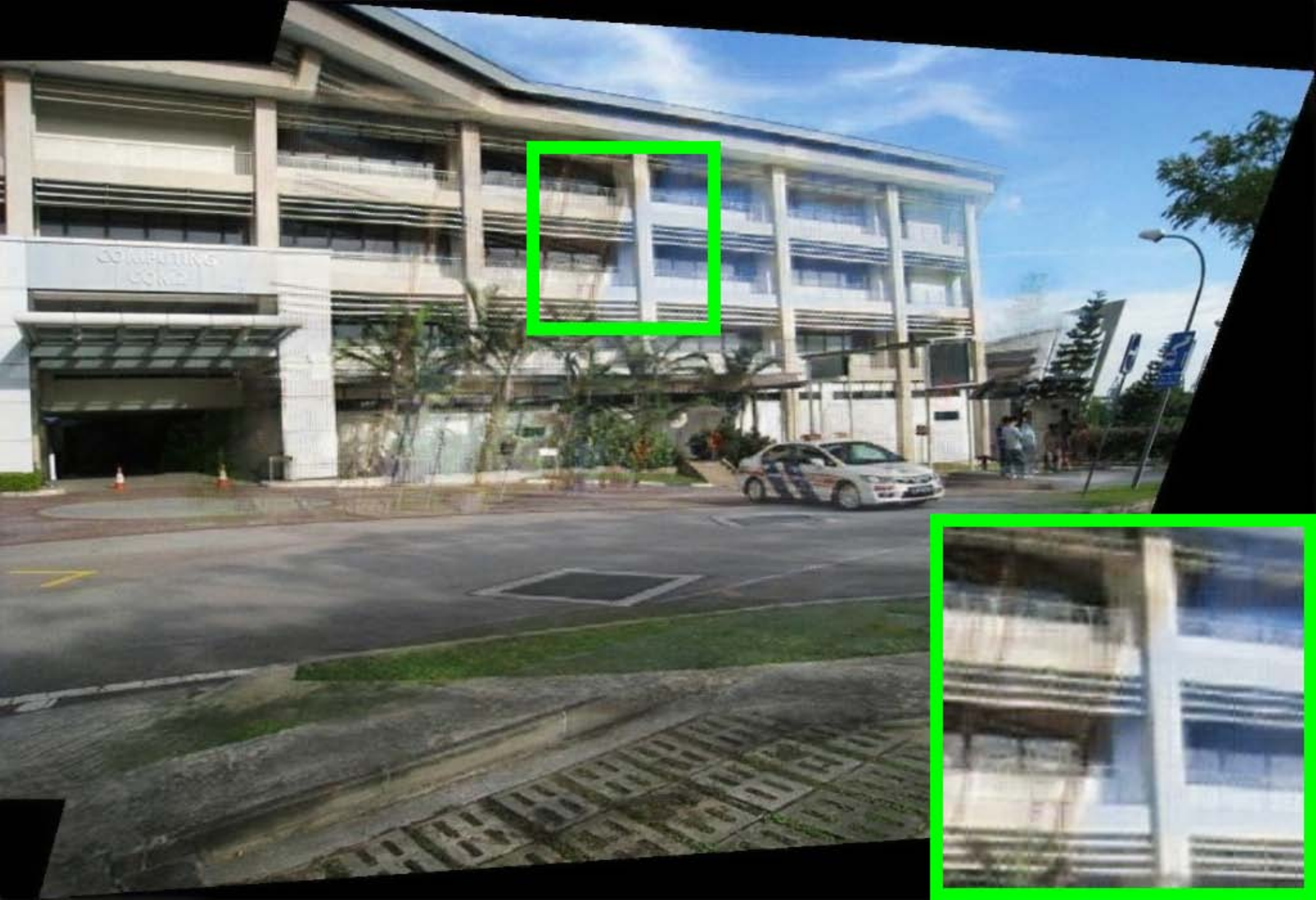}
		&\includegraphics[width=0.13\textwidth,height=0.06\textheight]{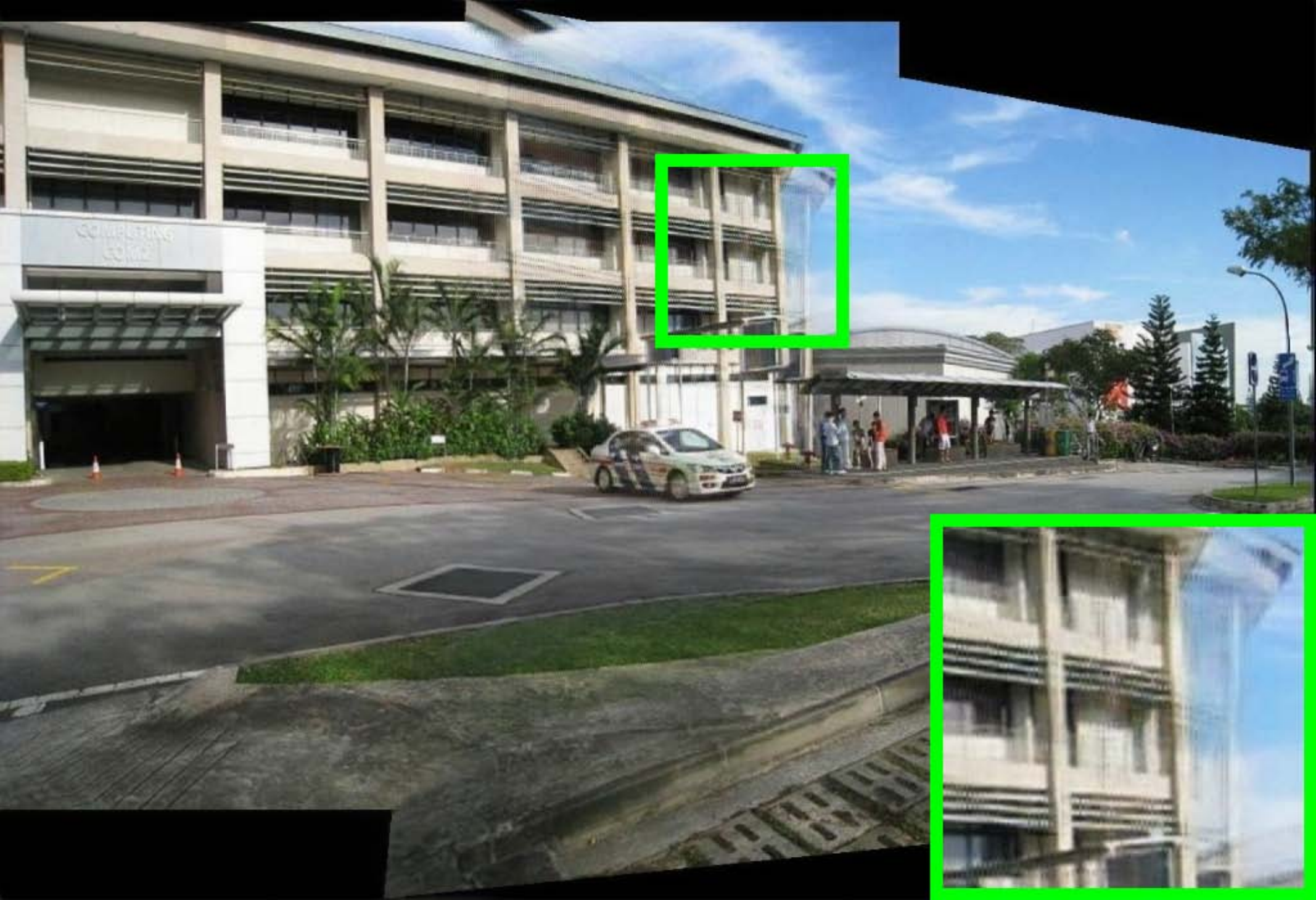}
		&\includegraphics[width=0.13\textwidth,height=0.06\textheight]{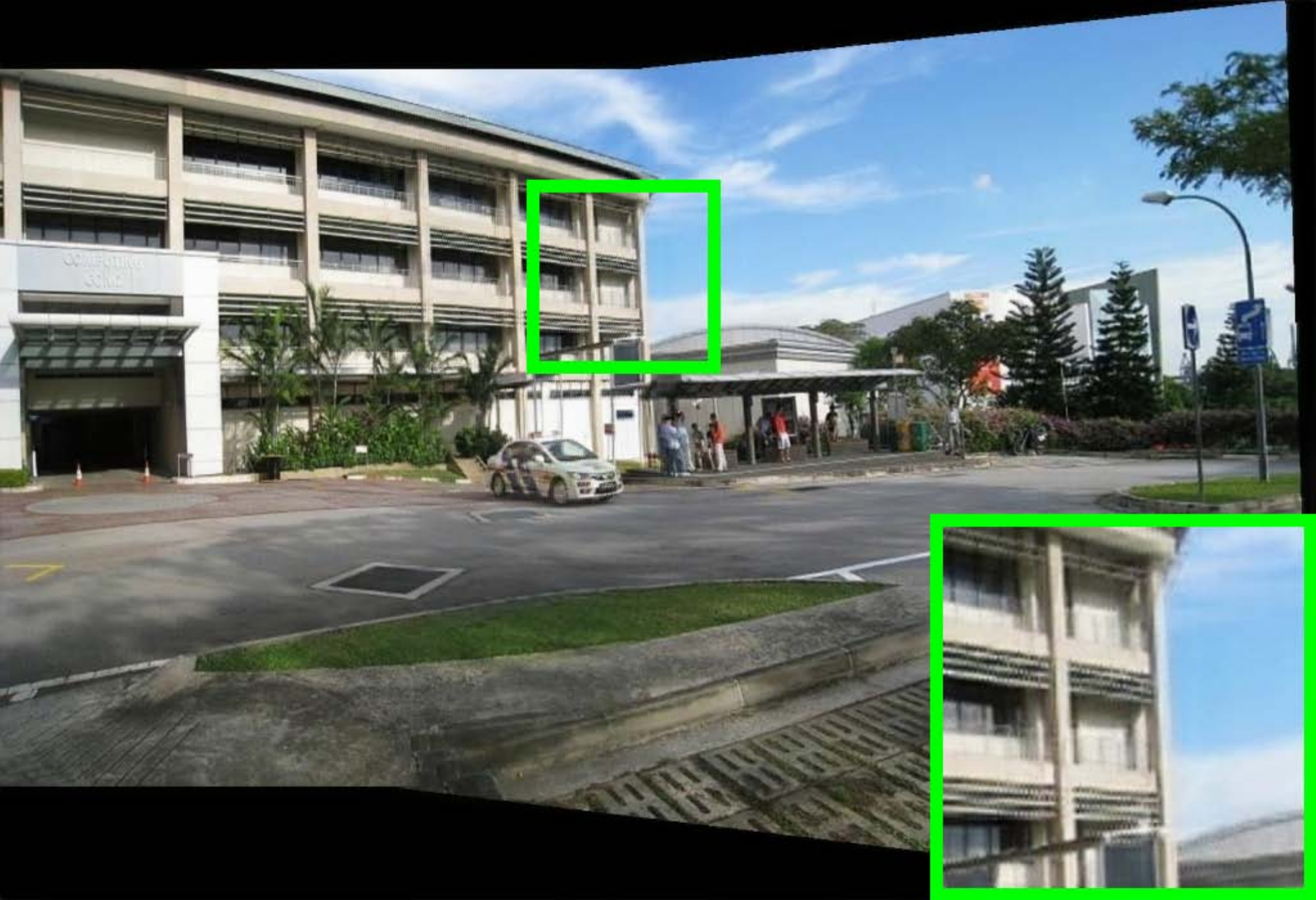}\tabularnewline
		&Input&VFIS&RSFI&Ours&&&Input&VFIS&RSFI&Ours\\
	\end{tabular}
	\caption{Visual comparisons of different adversarially trained stitching models on benign and attacked images.}
	\label{fig:exp2}
\end{figure*}
\begin{table*}[!h]
	\begin{center}
		\footnotesize
		\setlength
		\tabcolsep{3pt}
		\centering
		\begin{tabular}{p{0.7cm}<{\centering}|p{1.4cm}<{\raggedright}|p{1.4cm}<{\raggedright}|p{1.4cm}<{\centering}|p{1.4cm}<{\centering}|p{1.4cm}<{\centering}|p{1.4cm}<{\centering}|p{1.4cm}<{\centering}|p{1.4cm}<{\centering}|p{1.4cm}<{\centering}|p{1.4cm}<{\centering}}\hline	
			Data  & Method & Type  & EN~$\uparrow$   & SF~$\uparrow$    & SD~$\uparrow$    & AG~$\uparrow$    & SIQE~$\uparrow$  & BR~$\uparrow$    & NIQE~$\downarrow$  & PI~$\downarrow$  \\\hline
			\multirow{6}{*}{\rotatebox{90}{UDIS-D}} & \multicolumn{1}{l|}{\multirow{2}{*}{VFIS}} & Benign & 7.280 & 17.091 & 57.228 & 6.891 & 41.101 & 34.621 & 4.043 & 3.041 \\
			&       & Attacked &\cellcolor{gray!20} 6.928 &\cellcolor{gray!20} 16.300 &\cellcolor{gray!20} 53.403 & \cellcolor{gray!20}6.192 & \cellcolor{gray!20}30.186 & \cellcolor{gray!20}19.773 &\cellcolor{gray!20} 4.731 & \cellcolor{gray!20}3.816 \\
			\cline{2-11}          & \multicolumn{1}{l|}{\multirow{2}{*}{RSFI}} & Benign & 7.298 & 17.130 & 56.684 & 6.879 & 40.383 & 34.331 & 4.028 & 2.974 \\
			&       & Attacked & \cellcolor{gray!20} 7.011 & \cellcolor{gray!20} 16.511 &\cellcolor{gray!20}  54.525 &\cellcolor{gray!20}  6.204 & \cellcolor{gray!20} 31.788 & \cellcolor{gray!20} 20.158 &\cellcolor{gray!20}  4.653 & \cellcolor{gray!20} 3.692 \\
			\cline{2-11}          & \multicolumn{1}{l|}{\multirow{2}{*}{Ours}} & Benign &\textbf{7.353} & \textbf{17.320} & {56.731} & {7.258} &{47.675} &\textbf{34.502} & \textbf{3.904} &{2.945} \\
			&       & Attacked &\cellcolor{gray!20}  7.129 & \cellcolor{gray!20} 16.847 & \cellcolor{gray!20} \textbf{56.778} &\cellcolor{gray!20}  \textbf{7.631} & \cellcolor{gray!20} \textbf{47.963} & \cellcolor{gray!20} 31.426 & \cellcolor{gray!20} 4.281 & \cellcolor{gray!20} \textbf{2.832} \\
			\hline\hline
			\multirow{6}{*}{\rotatebox{90}{RWCC}} & \multicolumn{1}{l|}{\multirow{2}{*}{VFIS}} & Benign & 7.253 & 13.929 & 63.158 & 5.925 & 39.108 & 27.836 & 3.678 & 3.743 \\
			&       & Attacked & \cellcolor{gray!20} 6.728 &\cellcolor{gray!20}  12.339 & \cellcolor{gray!20} 56.945 & \cellcolor{gray!20} 4.879 & \cellcolor{gray!20} 21.741 &\cellcolor{gray!20}  23.129 &\cellcolor{gray!20}  4.388 & \cellcolor{gray!20} 4.448 \\
			\cline{2-11}          & \multicolumn{1}{l|}{\multirow{2}{*}{RSFI}} & Benign & 7.274 & 13.856 & 63.027 & 5.979 & 39.032 & 27.365 & 3.689 & 3.960 \\
			&       & Attacked & \cellcolor{gray!20} 6.817 & \cellcolor{gray!20} 12.447 &\cellcolor{gray!20}  56.794 &\cellcolor{gray!20}  5.001 &\cellcolor{gray!20}  23.328 & \cellcolor{gray!20} 23.466 & \cellcolor{gray!20} 4.253 &\cellcolor{gray!20}  4.325 \\
			\cline{2-11}          & \multicolumn{1}{l|}{\multirow{2}{*}{Ours}} & Benign & \textbf{8.327} & \textbf{14.276} & {64.280} & \textbf{6.217} & {39.490} & \textbf{28.396} & {3.611} &{3.699} \\
			&       & Attacked & \cellcolor{gray!20} 7.923 & \cellcolor{gray!20} 13.438 &\cellcolor{gray!20}  \textbf{65.932} & \cellcolor{gray!20} 5.732 & \cellcolor{gray!20} \textbf{40.117} &\cellcolor{gray!20}  27.357 & \cellcolor{gray!20} \textbf{3.526} & \cellcolor{gray!20} \textbf{3.454} \\\hline
		\end{tabular}
		\caption{Quantitative comparison between adversarially trained models and the proposed method on benign and attacked data.}
		%		\vspace{-0.5em}		
		\label{tab:exp2}
	\end{center}
	%\vspace{-1em}
\end{table*}
\subsection{Implementation Details}
There are two benchmarks available for image stitching, including a synthetic dataset based on MS-COCO~\cite{nie2020view} and a real-world UDIS-D~\cite{nie2021unsupervised} collected from various moving videos. For the training of homography estimation module, we employed the synthesized MS-COCO dataset for initial~$120$ epochs and fine-tuned on the training set of UDIS-D for~$20$ epochs. The optimizer is Adam~\cite{kingma2014adam} with an initial learning rate of~$1e^{-4}$ and the decay rate is~$0.96$. The reconstruction module is trained on UDIS-D for~$30$ epoch, adhering to the same hyperparameter configuration. For evaluation, the test set of UDIS-D is adopted, which contains~$1106$ image pairs. Moreover, we additionally obtained~$62$ pairs of real-world challenging cases~(RWCC) from~\cite{zhang2020content,lin2015adaptive,chang2014shape,gao2011constructing,chen2016natural,li2017parallax} as comprehensive validation. For adversarial attack, perturbation intensity~$\epsilon$ is set as~$8/255$, the iteration count is~$3$, and the step size is~$5/255$. Both the training and testing are implemented on Pytorch with an NVIDIA Tesla A40 GPU.

\subsection{Performance on Image Stitching}

\begin{figure*}[h]
	\centering
	\includegraphics[width=0.9\textwidth]{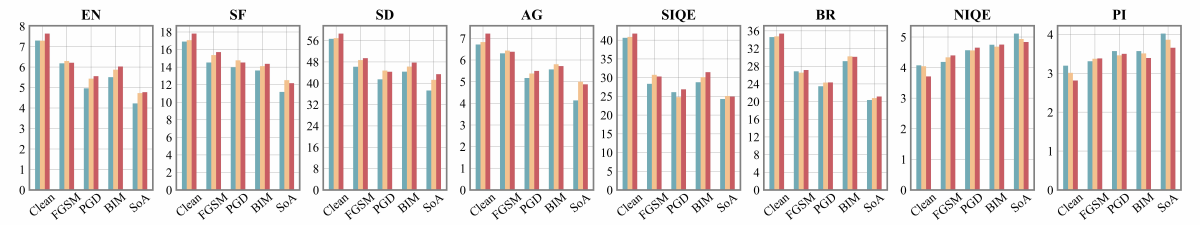}
%	\vspace{-1.5em}
	\caption{Performance deterioration from different attacks on deep learning models. Blue, yellow and red denote VFIS, RSFI and our baseline model trained with clean data.}
%	\vspace{-.5em}
	\label{fig:11}
\end{figure*}

\begin{figure*}[!h]
	\centering
	\setlength{\tabcolsep}{1pt}
	\begin{tabular}{cccccccccc}	
		\rotatebox{90}{\ \ \  Benign}&\includegraphics[width=0.07\textwidth,,height=0.07\textheight]{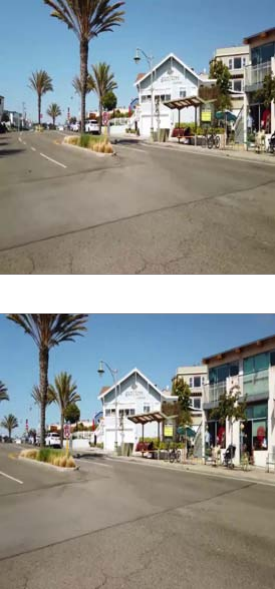}
		&\includegraphics[width=0.14\textwidth,height=0.07\textheight]{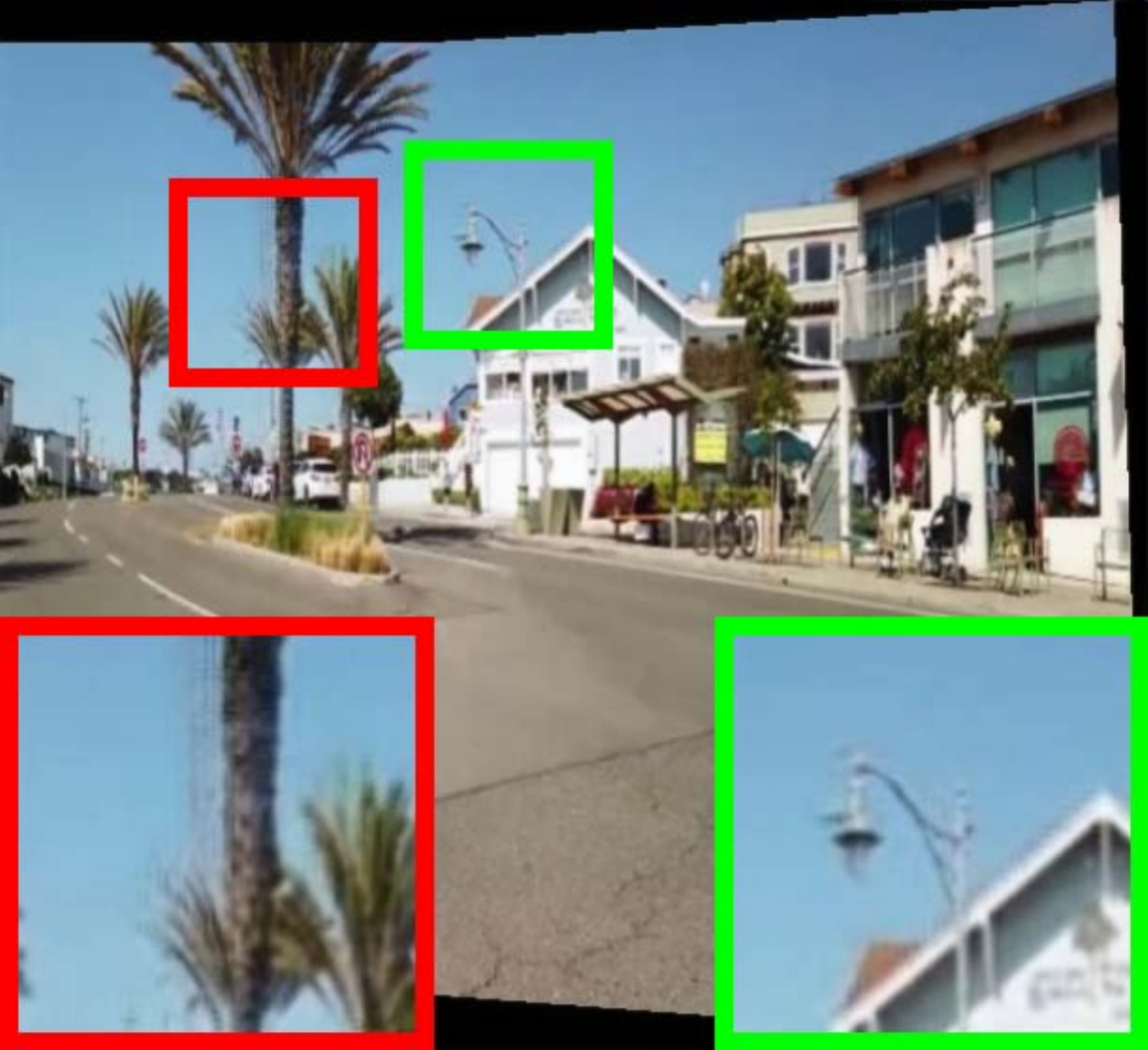}
		&\includegraphics[width=0.14\textwidth,height=0.07\textheight]{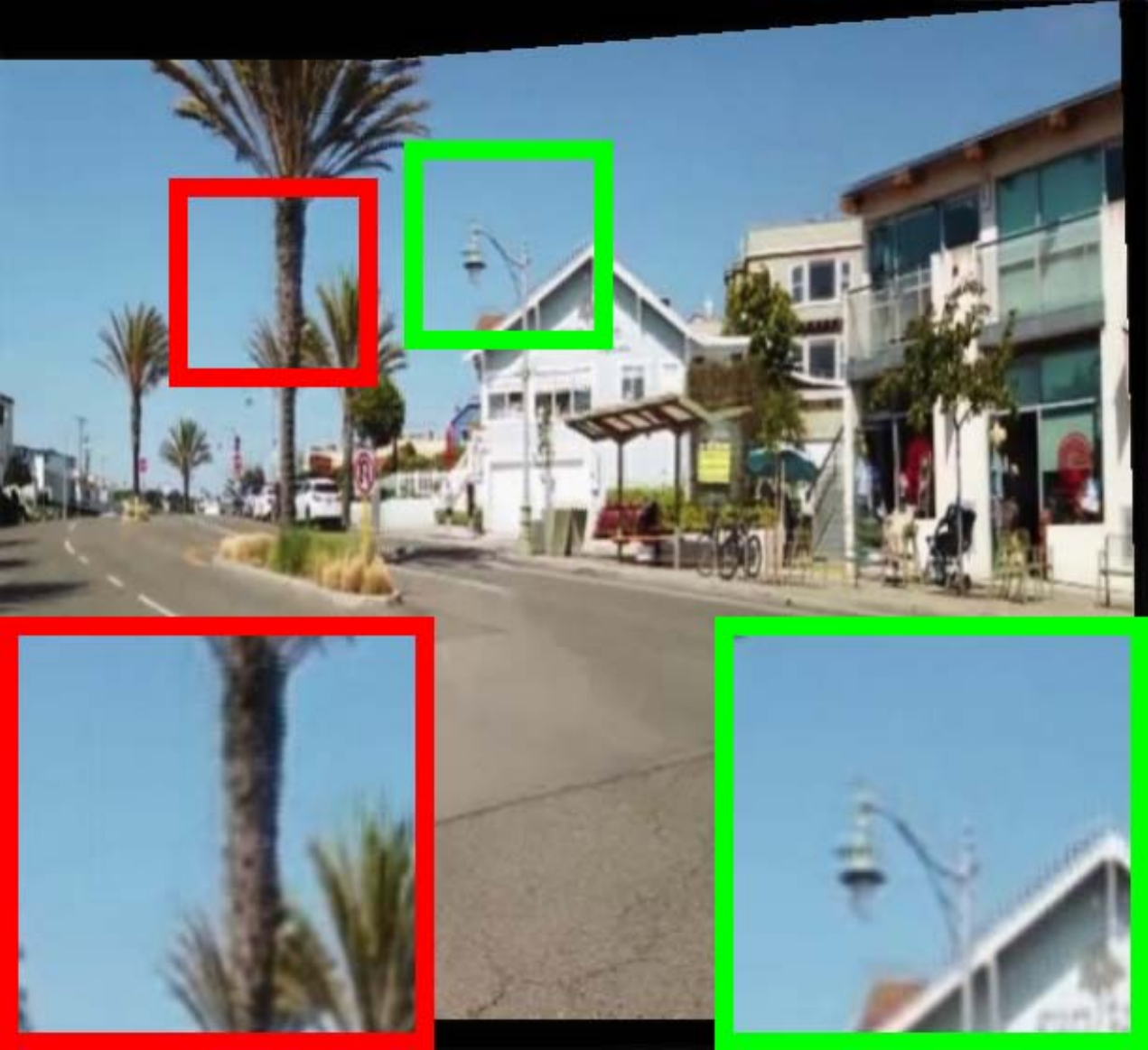}
		&\includegraphics[width=0.14\textwidth,height=0.07\textheight]{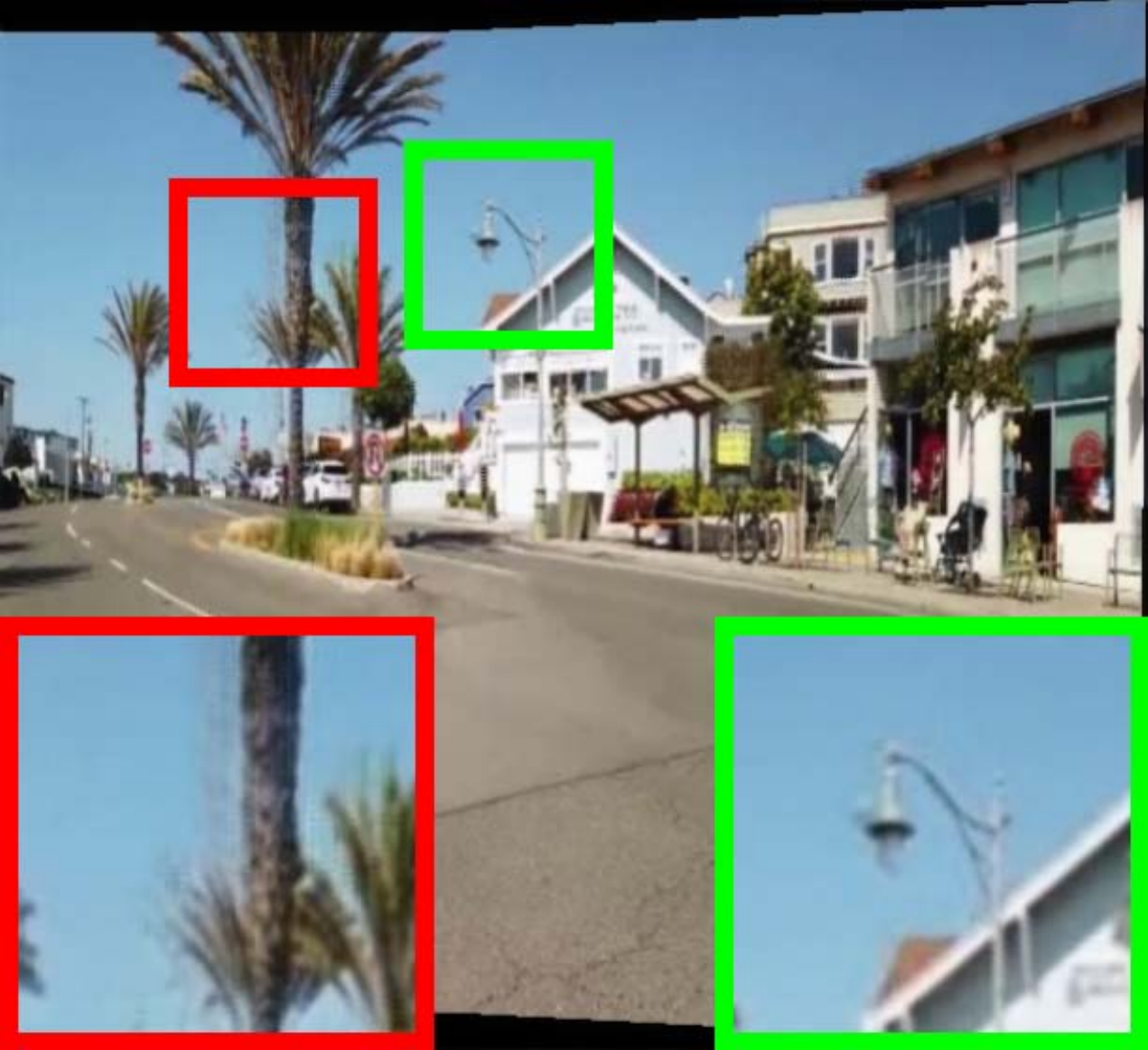}
		&\includegraphics[width=0.14\textwidth,height=0.07\textheight]{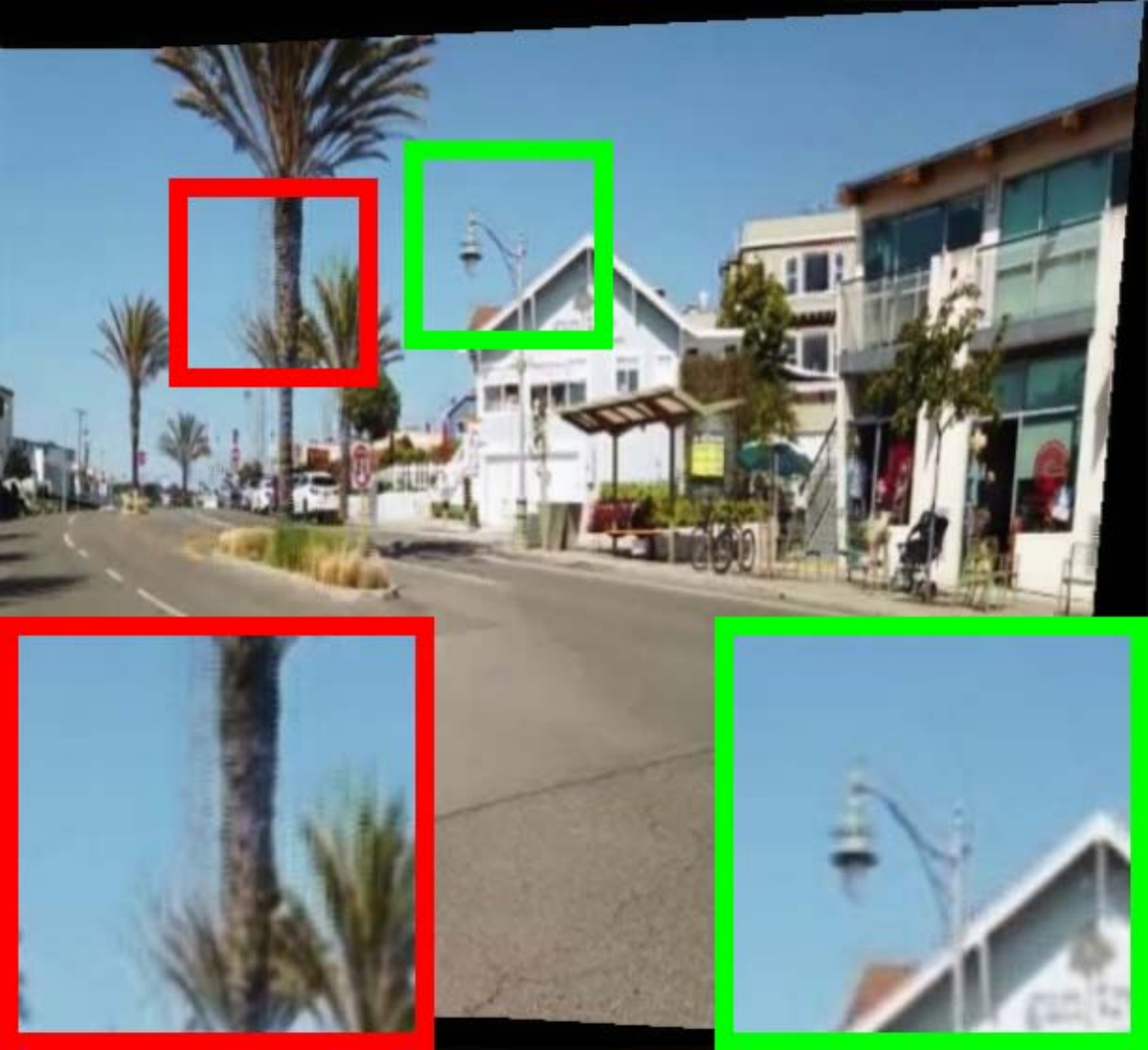}
		&\includegraphics[width=0.14\textwidth,height=0.07\textheight]{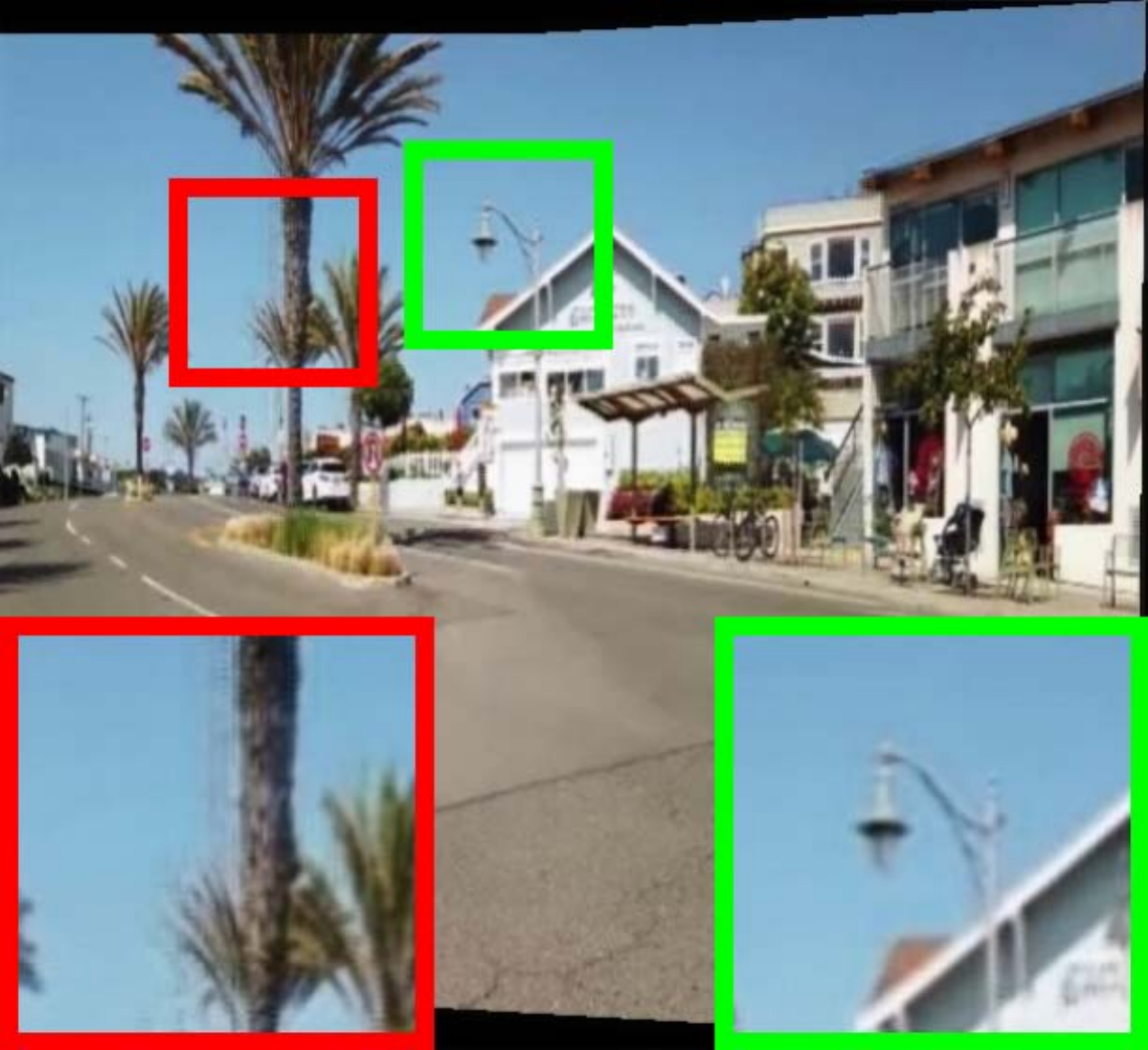}
		&\includegraphics[width=0.14\textwidth,height=0.07\textheight]{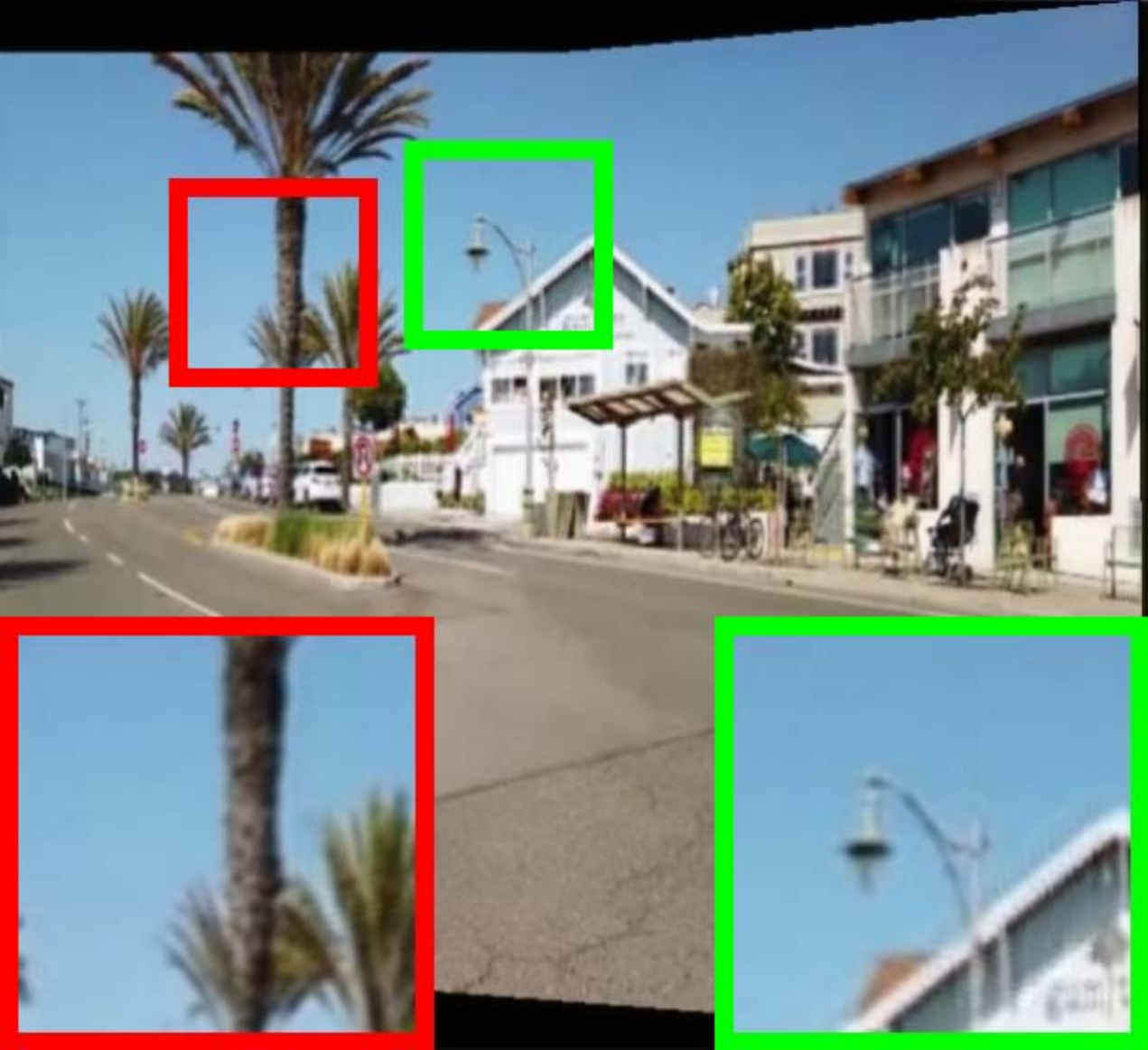}\tabularnewline
		\rotatebox{90}{\  \ Attacked}&\includegraphics[width=0.07\textwidth,height=0.07\textheight]{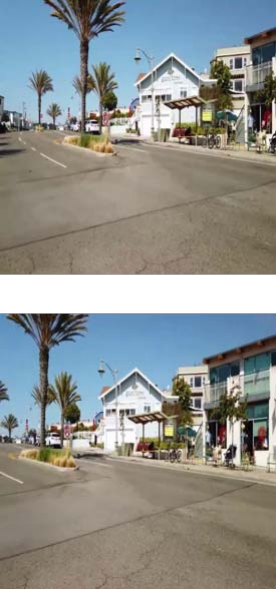}
		&\includegraphics[width=0.14\textwidth,height=0.07\textheight]{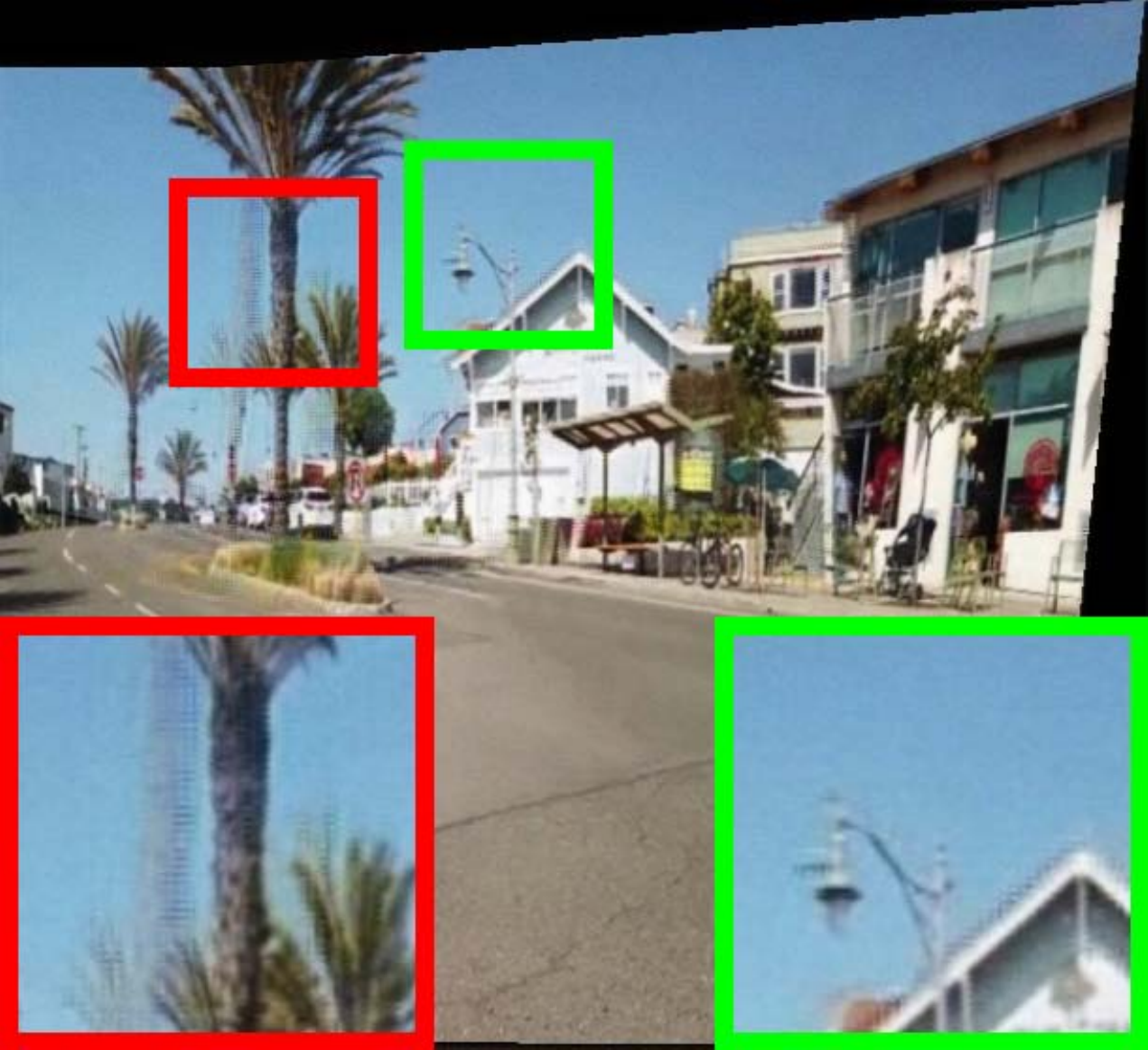}
		&\includegraphics[width=0.14\textwidth,height=0.07\textheight]{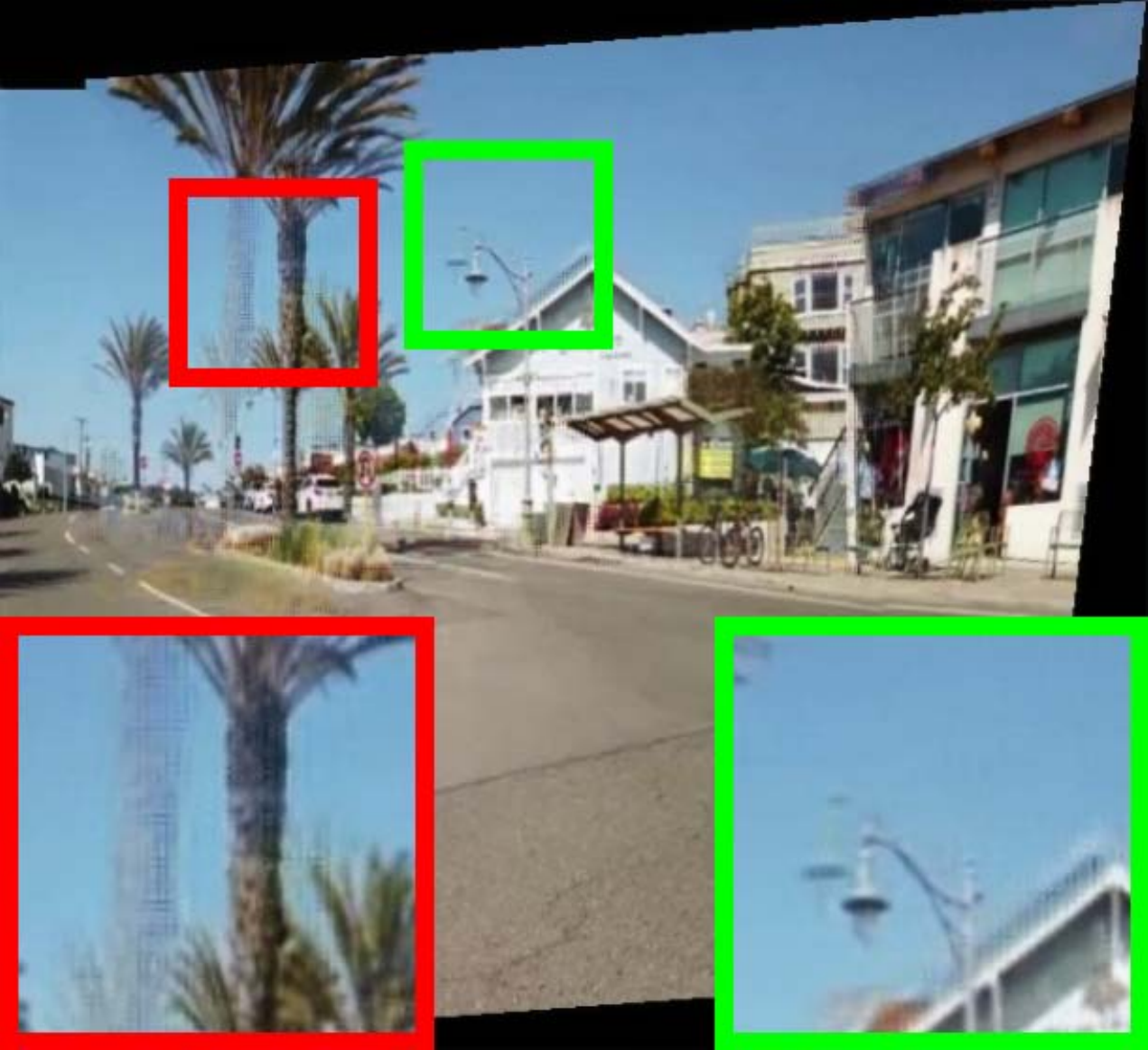}
		&\includegraphics[width=0.14\textwidth,height=0.07\textheight]{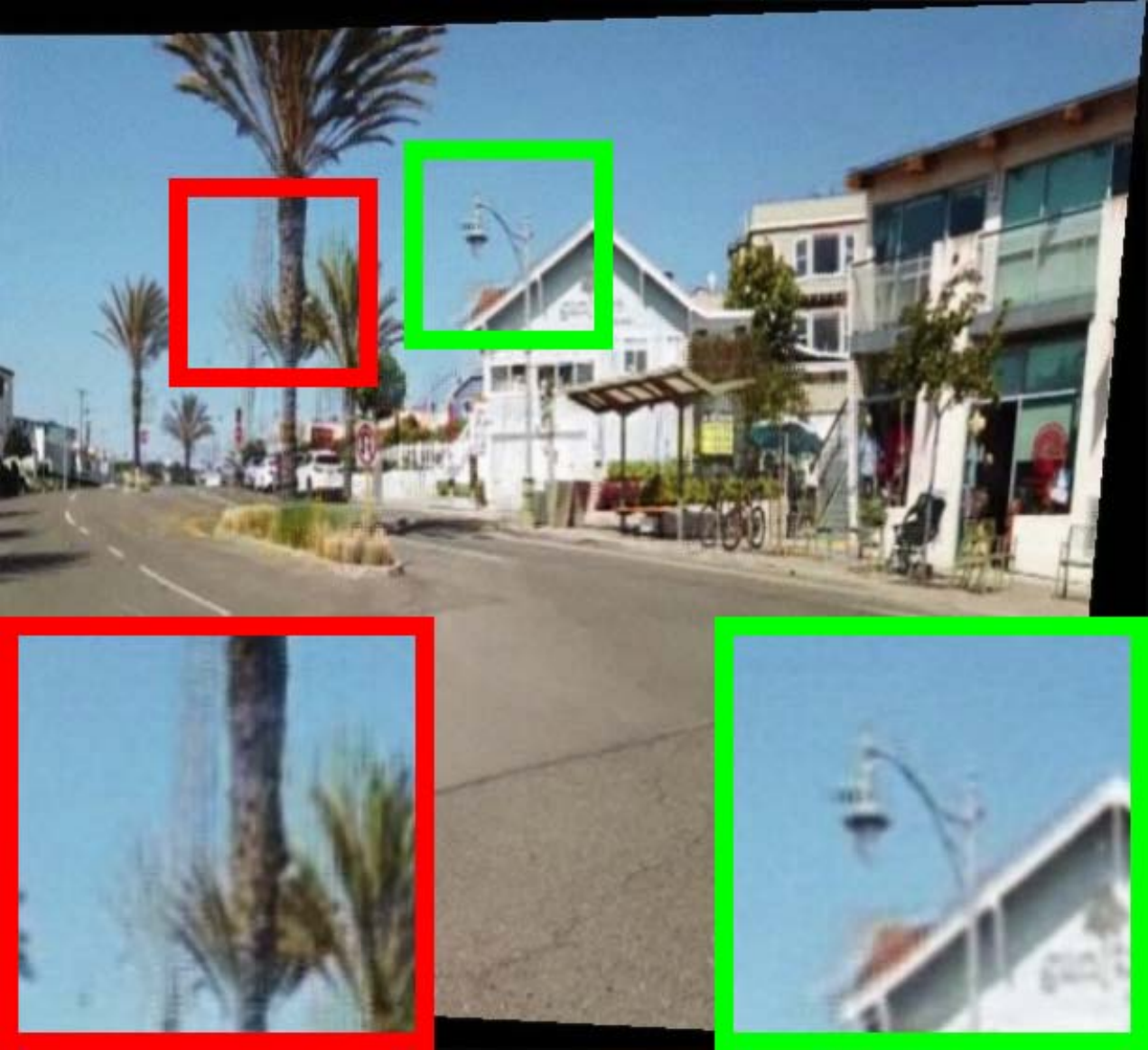}
		&\includegraphics[width=0.14\textwidth,height=0.07\textheight]{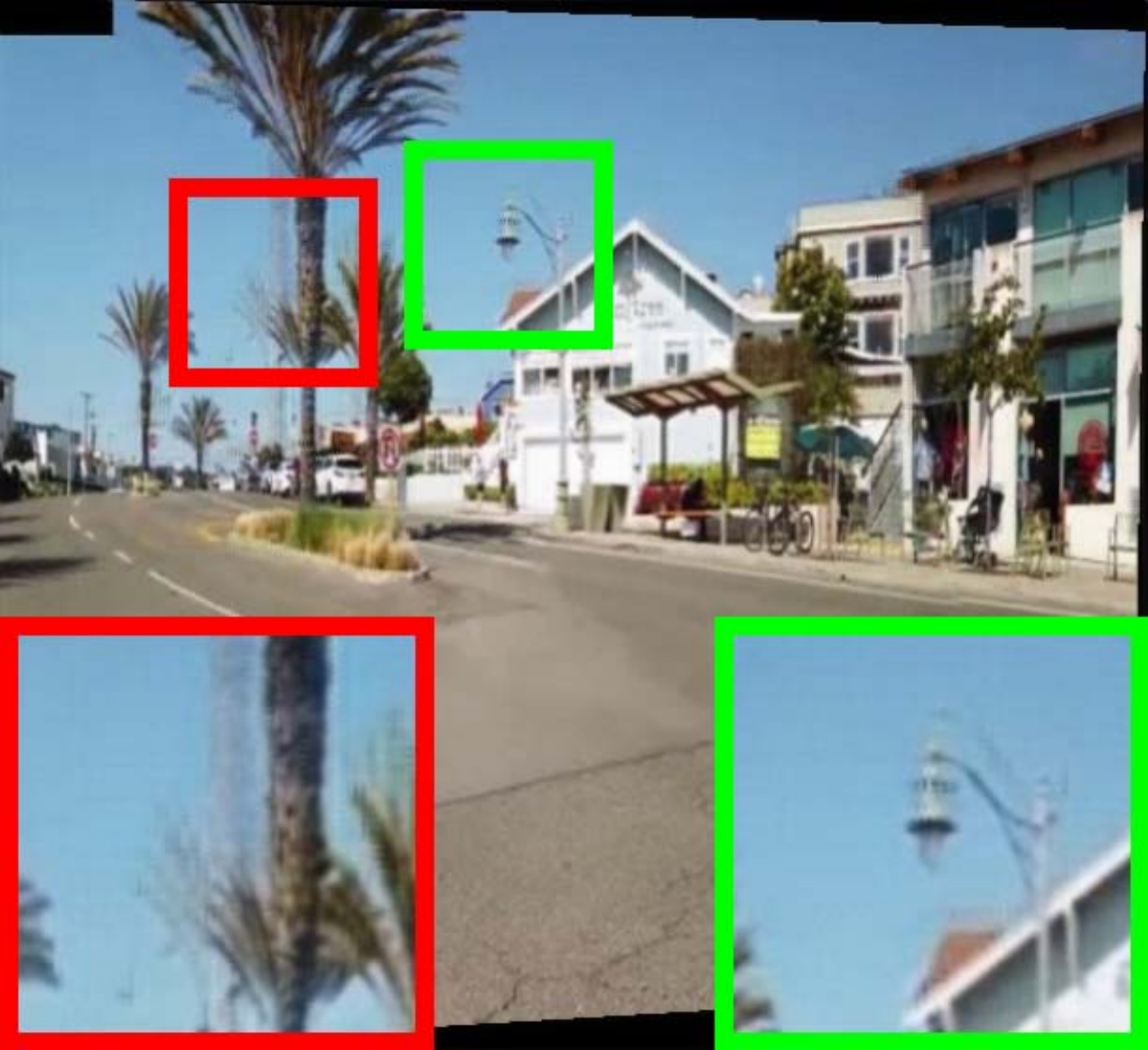}
		&\includegraphics[width=0.14\textwidth,height=0.07\textheight]{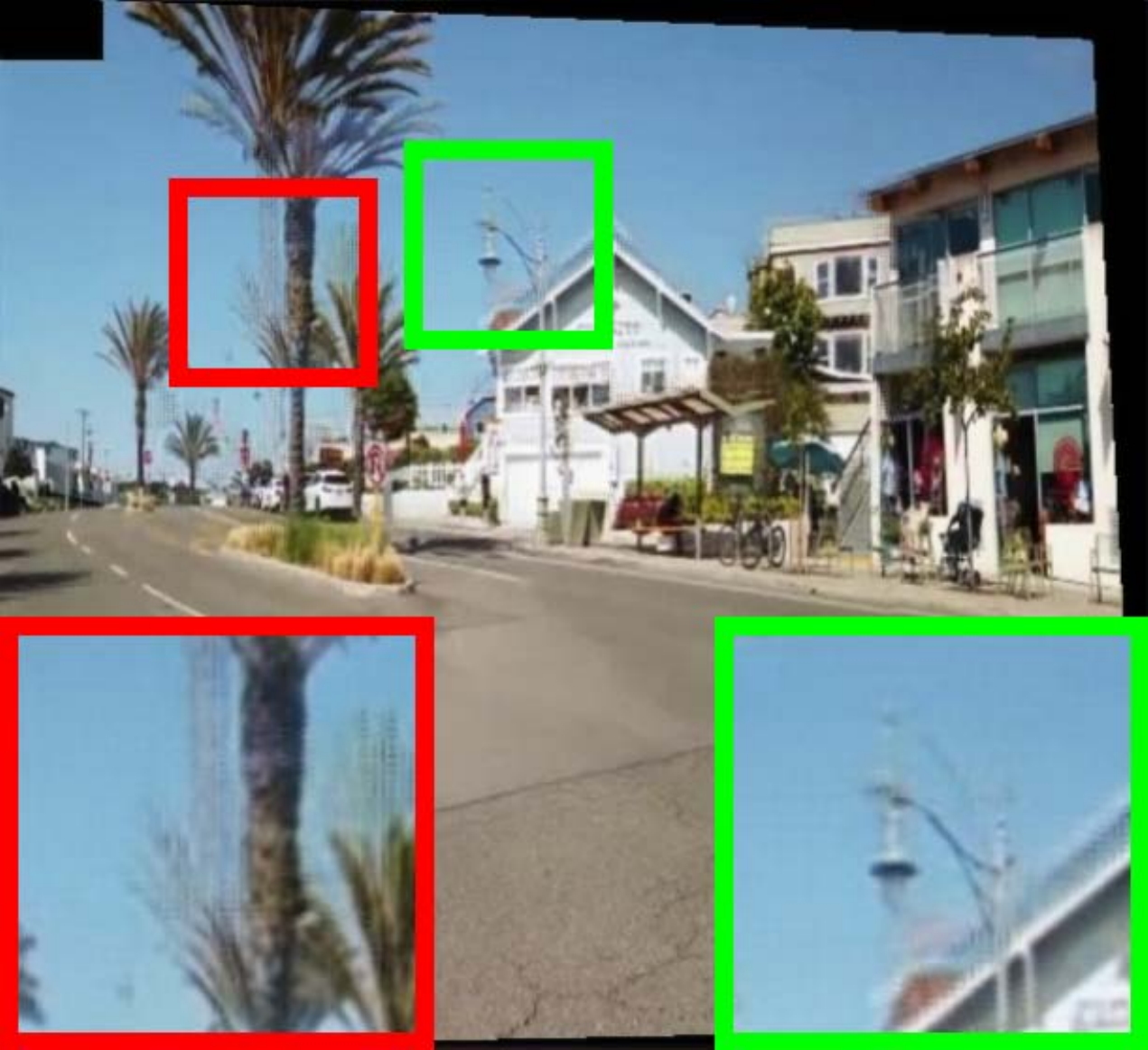}
		&\includegraphics[width=0.14\textwidth,height=0.07\textheight]{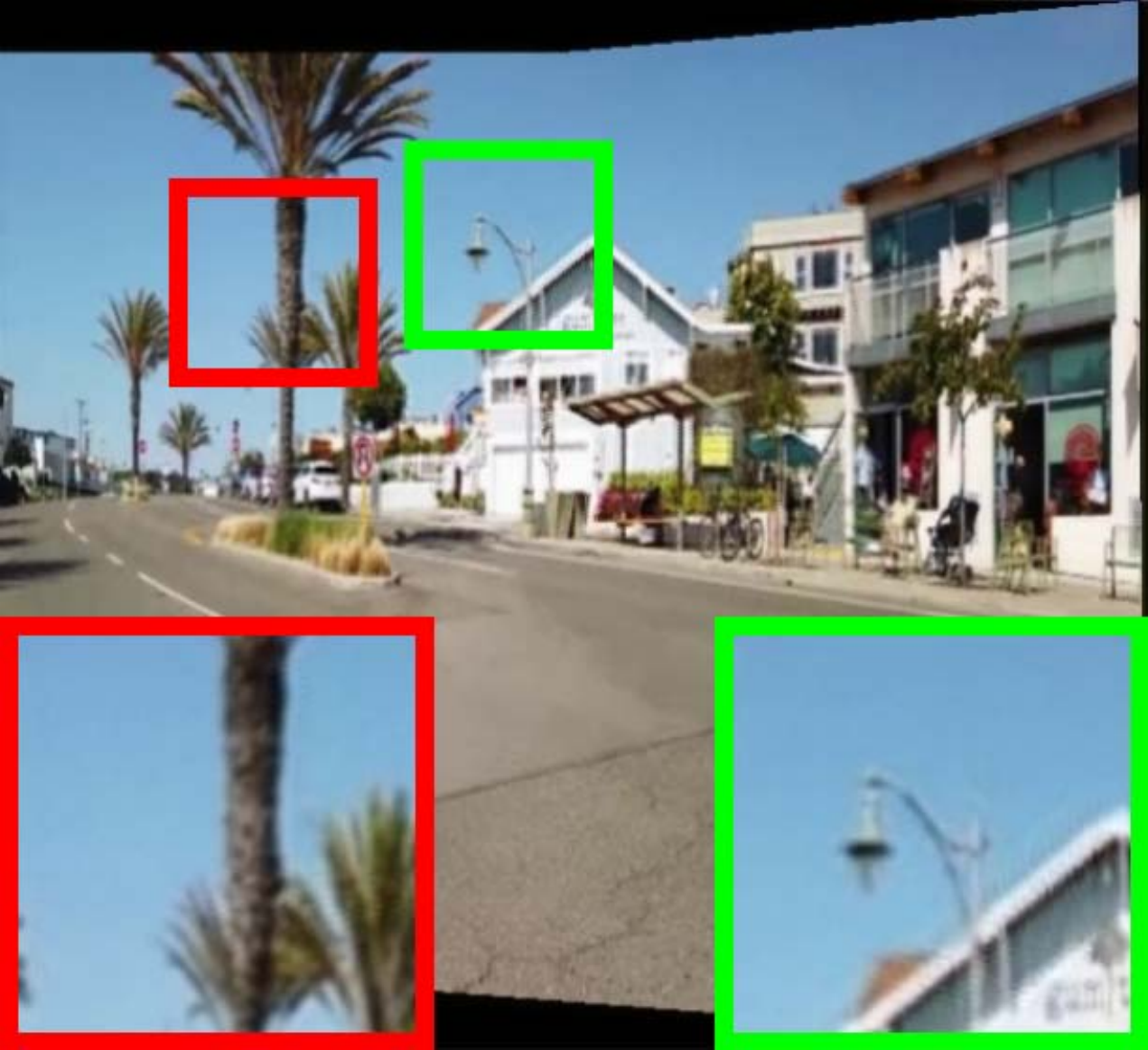}\tabularnewline
		&Input&Clean+AAT&FGSM+AAT&BIM+AAT&PGD+AAT&Hybrid+AAT&SoA+AAT(Ours)\\
	\end{tabular}
	%	\vspace{-.5em}
	\caption{Performance comparison of the proposed adaptive adversarial training~(AAT) mechanism incorporated with variant perturbations. Obviously, SoA + AAT strategy exhibits superior performance on both benign and attacked data.}
	%	\vspace{-1em}
	\label{fig:exp6}
\end{figure*}
\subsubsection{Evaluation on the compatibility against different attacks.}
We evaluated the stitching performance of our method against attacks from FGSM, BIM, PGD, and SoA strategies, as well as in benign (attack-free) scenarios. As illustrated in Fig.~\ref{fig:exp7}, the first sample is drawn from the UDIS-D dataset, while the second originates from the RWCC dataset. Notably, the bottom corners of the attacked scenarios provide visual representations of various attack perturbations. Our method clearly exhibits robust performance against SoA attacks and also showcases strong compatibility with FGSM, BIM, and PGD attacks. Moreover, in benign scenarios, our method maintains consistent performance, mitigating the degradation observed in routine adversarial training.
\subsubsection{Evaluation of AAT against routine adversarial training.}
Visual comparisons between the proposed adaptive adversarial training~(AAT) with the routine adversarial training are presented in Fig.~\ref{fig:exp2}. For these comparisons, we adopt the prolific deep learning based stitching methods VFIS~\cite{nie2020view} and RSFI~\cite{nie2021unsupervised}. Both methods were retrained using the projected gradient descent~\cite{madry2017towards} to derive their respective robust models. In the first sample, the results of VFIS and RSFI on benign images present the ghost effect within the red frame, while they illustrate misalignment on the attacked images within the green frame. In the second sample, VFIS exhibits misalignment in both benign and attacked images, and simultaneously introduces ghosting artifacts. In contrast, the proposed method consistently delivers reliable stitching results under both attacked and benign conditions. 

Quantitative results on UDIS-D and RWCC are shown in Table.~\ref{tab:exp2}, where Entropy~(EN)~\cite{zhao2023cddfuse}, Spatial Frequency~(SF)~\cite{eskicioglu1995image}, Standard Deviation~(SD)~\cite{zhao2022discrete}, Average Gradient~(AG)~\cite{cui2015detail}, Stitched Image Quality Evaluator~(SIQE)~\cite{madhusudana2019subjective}, Blind/Referenceless image spatial quality evaluator~(BR)~\cite{mittal2012no}, Naturalness Image Quality Evaluator~\cite{mittal2012no}~(NIQE) and Perceptual Index~(PI) are employed as metrics. For the first six metrics, higher values mean better image quality; for the last two, lower values are preferable. After adversarial training, both VFIS and RSFI exhibit decreased performance compared to their results on benign data. Although adversarial training bolsters the robustness of these models, there remains a discernible performance gap compared to the benign counterparts. Our method consistently exhibits minimal disparity in performance between the attacked and benign versions.
%, striking a balance between robustness and efficiency.

\begin{table*}[h]
	\begin{center}
		\footnotesize
		\setlength\tabcolsep{3pt}
		\centering
		\begin{tabular}{p{0.7cm}<{\centering}|p{1.4cm}<{\raggedright}|p{1.4cm}<{\centering}|p{1.4cm}<{\centering}|p{1.4cm}<{\centering}|p{1.4cm}<{\centering}|p{1.4cm}<{\centering}|p{1.4cm}<{\centering}|p{1.4cm}<{\centering}|p{1.4cm}<{\centering}}\hline	Method & Type  & EN~$\uparrow$   & SF~$\uparrow$    & SD~$\uparrow$    & AG~$\uparrow$    & SIQE~$\uparrow$  & BR~$\uparrow$    & NIQE~$\downarrow$  & PI~$\downarrow$  \\
			\hline
			\multicolumn{1}{l|}{\multirow{2}[2]{*}{Bayesian}} & Benign & 7.230 & 17.094 & 57.594 & 7.083 & 45.023 & 32.442 & 4.293 & 3.090 \\
			& Attacked &\cellcolor{gray!20}  6.818 & \cellcolor{gray!20} 16.593 &\cellcolor{gray!20}  54.451 &\cellcolor{gray!20}  6.316 &\cellcolor{gray!20}  29.960 &\cellcolor{gray!20}  21.421 & \cellcolor{gray!20} 4.705 &\cellcolor{gray!20}  3.657 \\\hline\hline
			\multicolumn{1}{l|}{\multirow{2}[2]{*}{Quasi-Newton}} & Benign & 7.250 & 17.160 & 57.061 & 6.910 & 44.071 & 33.636 & 4.124 & 3.149 \\
			& Attacked & \cellcolor{gray!20} 6.996 & \cellcolor{gray!20} 16.633 &\cellcolor{gray!20}  54.174 &\cellcolor{gray!20}  6.353 &\cellcolor{gray!20}  28.073 &\cellcolor{gray!20}  22.198 &\cellcolor{gray!20}  4.648 & \cellcolor{gray!20} 3.722 \\\hline\hline
			\multicolumn{1}{l|}{\multirow{2}[1]{*}{Ours}} & Benign & \textbf{7.353} & \textbf{17.320} &{56.731} & {7.258} & {47.675} & \textbf{34.502} & \textbf{3.904} & {2.945} \\
			& Attacked & \cellcolor{gray!20} 7.129 &\cellcolor{gray!20}  16.847 &\cellcolor{gray!20}  \textbf{56.778} & \cellcolor{gray!20} \textbf{7.631} &\cellcolor{gray!20}  \textbf{47.963} & \cellcolor{gray!20} 31.426 &\cellcolor{gray!20}  4.281 &\cellcolor{gray!20}  \textbf{2.832} \\\hline
		\end{tabular}	
		\caption{Robustness comparison of the proposed AAT mechanism with different optimization solvers.}
%		\vspace{-1em}
		\label{tab:dataset_comparison}	
	\end{center}
\end{table*}

\subsection{Ablation Study}
\subsubsection{Analysis on the performance degradation.}

To assess the impact of the proposed SoA perturbations on stitching performance, we illustrate the performance degradation resulting from various attack strategies, including FGSM, PGD, BIM, and SoA, on deep learning based stitching models~(i.e., VFIS, RSFI, and our baseline model trained on benign data). For comparison, we also present the performance of these three models on clean images. In Fig.~\ref{fig:11}, blue, yellow, and red correspond to the VFIS, RSFI, and our baseline models, respectively. Clearly, the SoA perturbations result in a more significant performance degradation across all models when compared to other attack strategies. Based on these observations, we adopted SoA perturbations for adversarial training to bolster the robustness of stitching algorithms.

\subsubsection{Analysis on the proposed SoA.}

The proposed method incorporates SoA perturbations into the adaptive adversarial training~(AAT) to facilitate a robust stitching model. To further investigate the AAT mechanism, we combine it with different attack perturbations. A visual comparison is presented in Fig.~\ref{fig:exp6}, with 'Clean' denoting a training scenario without any attack perturbations, and 'Hybrid' perturbation emerges from amalgamating FGSM, BIM, and PGD perturbations. Notably, the model trained in Clean + AAT strategy reveals a significant disparity in performance between stitching benign images and their attacked counterparts. Models subjected to adversarial training using FGSM, BIM, or PGD perturbations demonstrate moderate robustness in the face of potent attacks, but this robustness is somewhat elevated when trained with Hybrid perturbations. Remarkably, our model stands out with unparalleled robustness,  maintaining high efficiency even with clean images.

\begin{figure}[h]
	\centering
	\setlength{\tabcolsep}{1pt}
	\begin{tabular}{cccccccccccc}	
		\rotatebox{90}{\ Benign}&\includegraphics[width=0.04\textwidth,height=0.055\textheight]{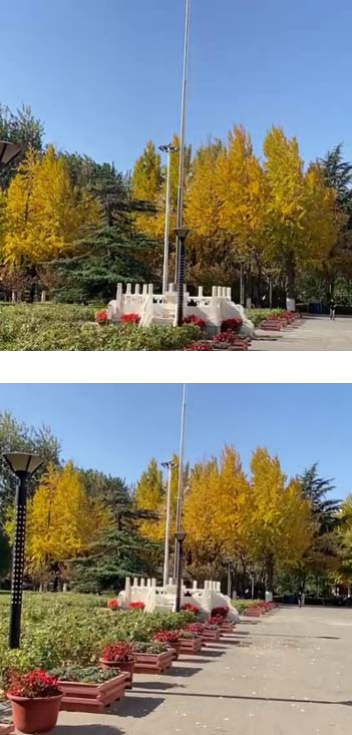}
		&\includegraphics[width=0.13\textwidth,height=0.055\textheight]{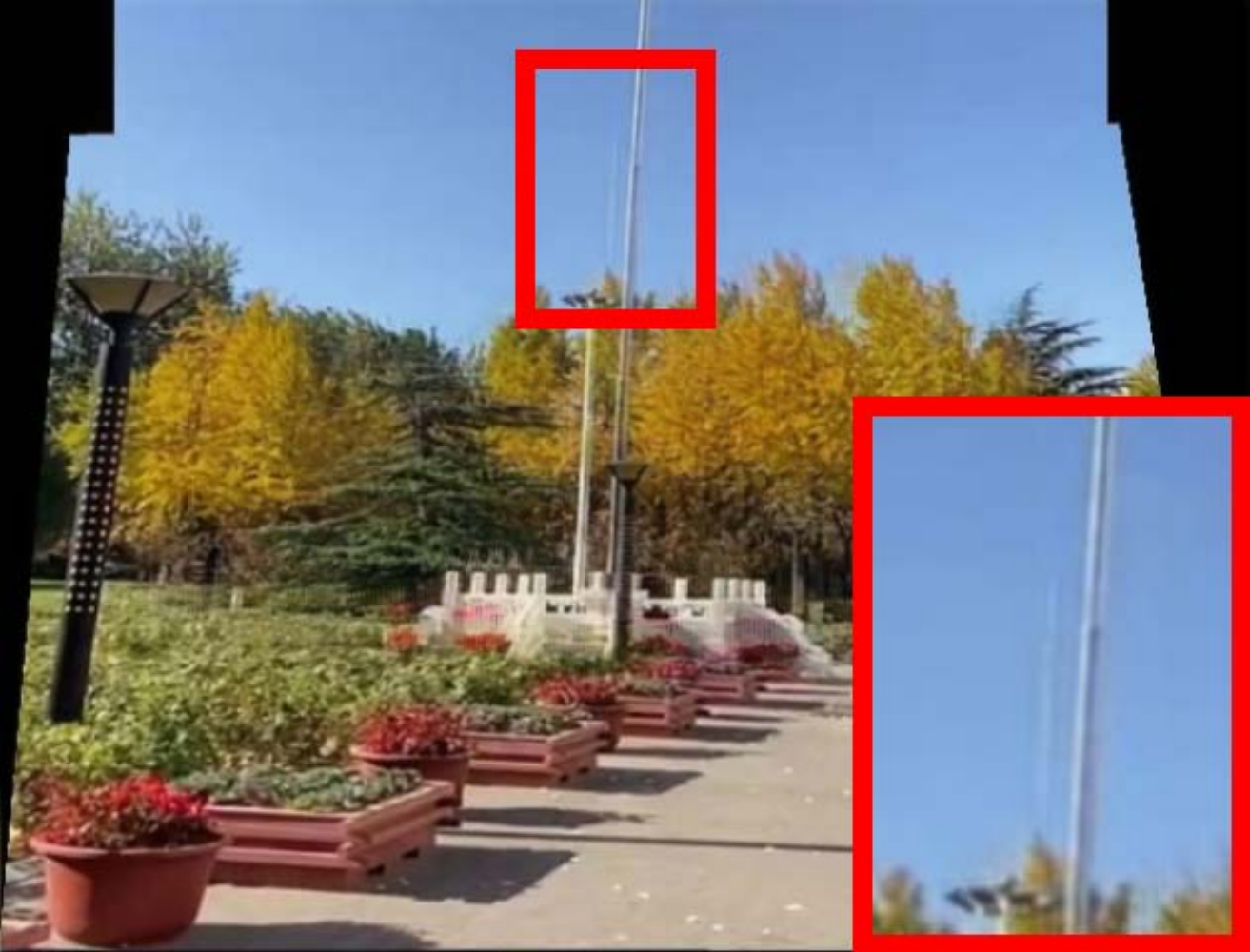}
		&\includegraphics[width=0.13\textwidth,height=0.055\textheight]{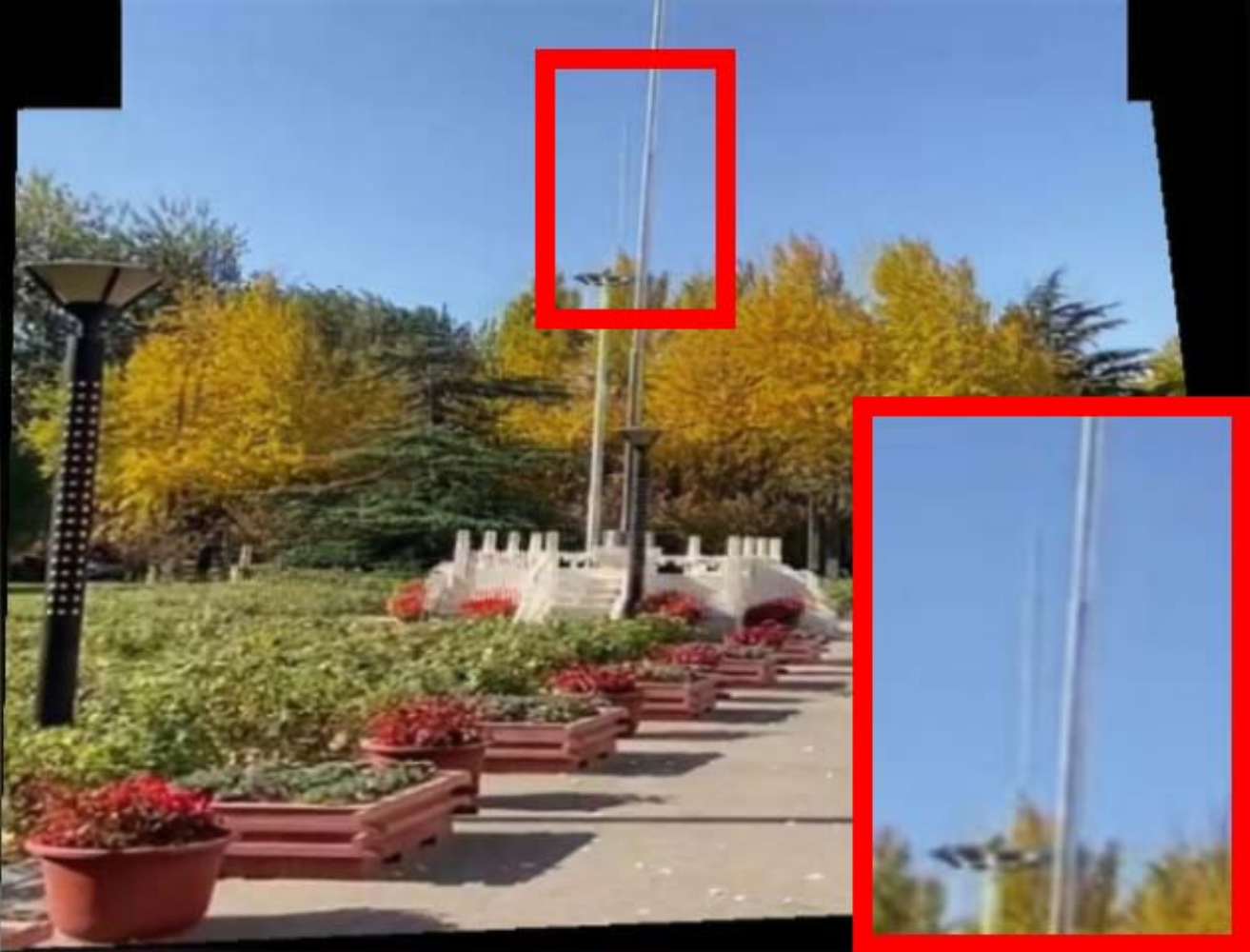}
		&\includegraphics[width=0.13\textwidth,height=0.055\textheight]{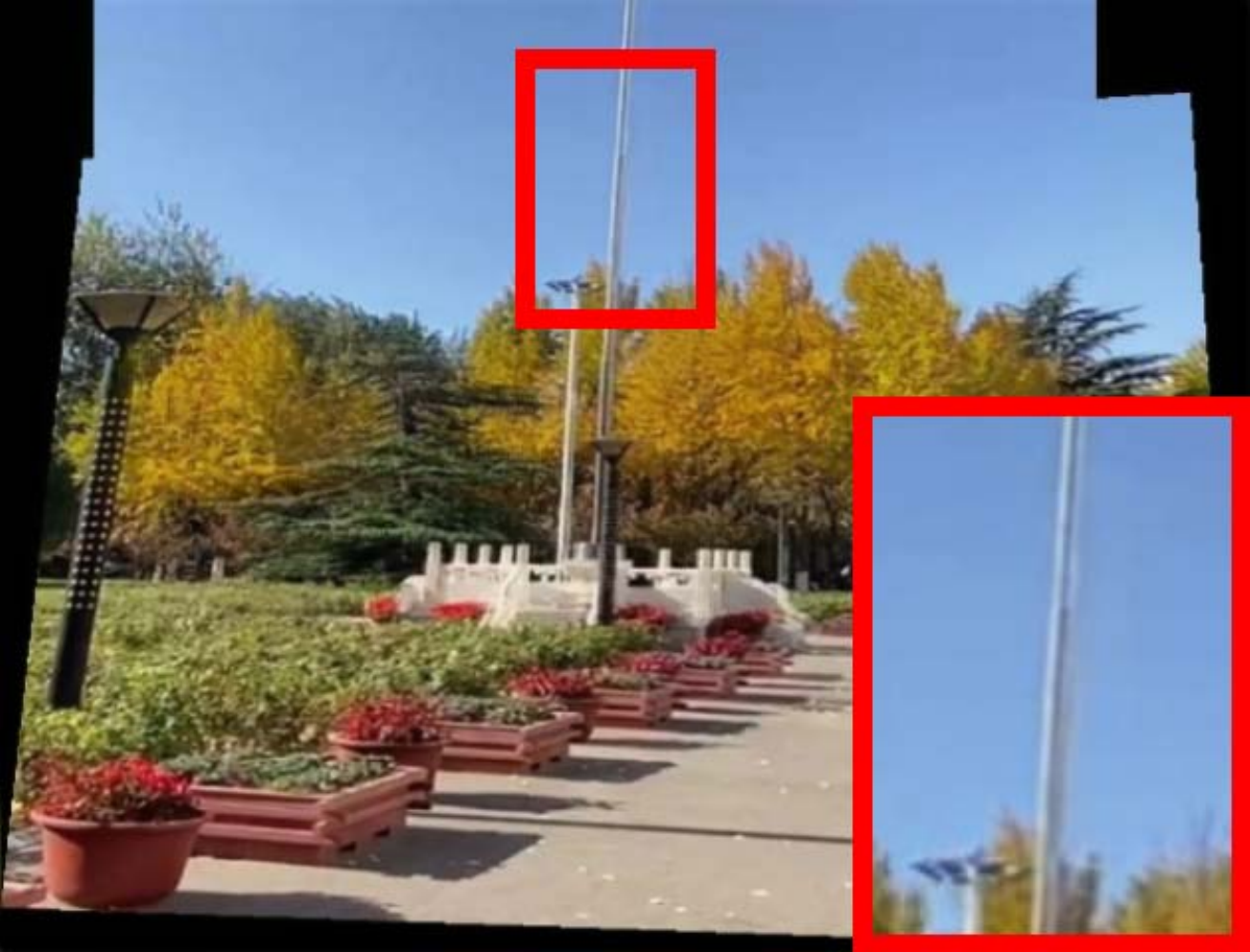}\\
		\rotatebox{90}{Attacked}&\includegraphics[width=0.04\textwidth,height=0.055\textheight]{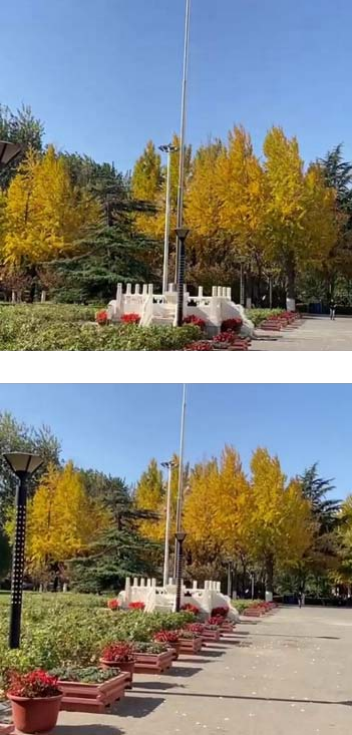}
		&\includegraphics[width=0.13\textwidth,height=0.055\textheight]{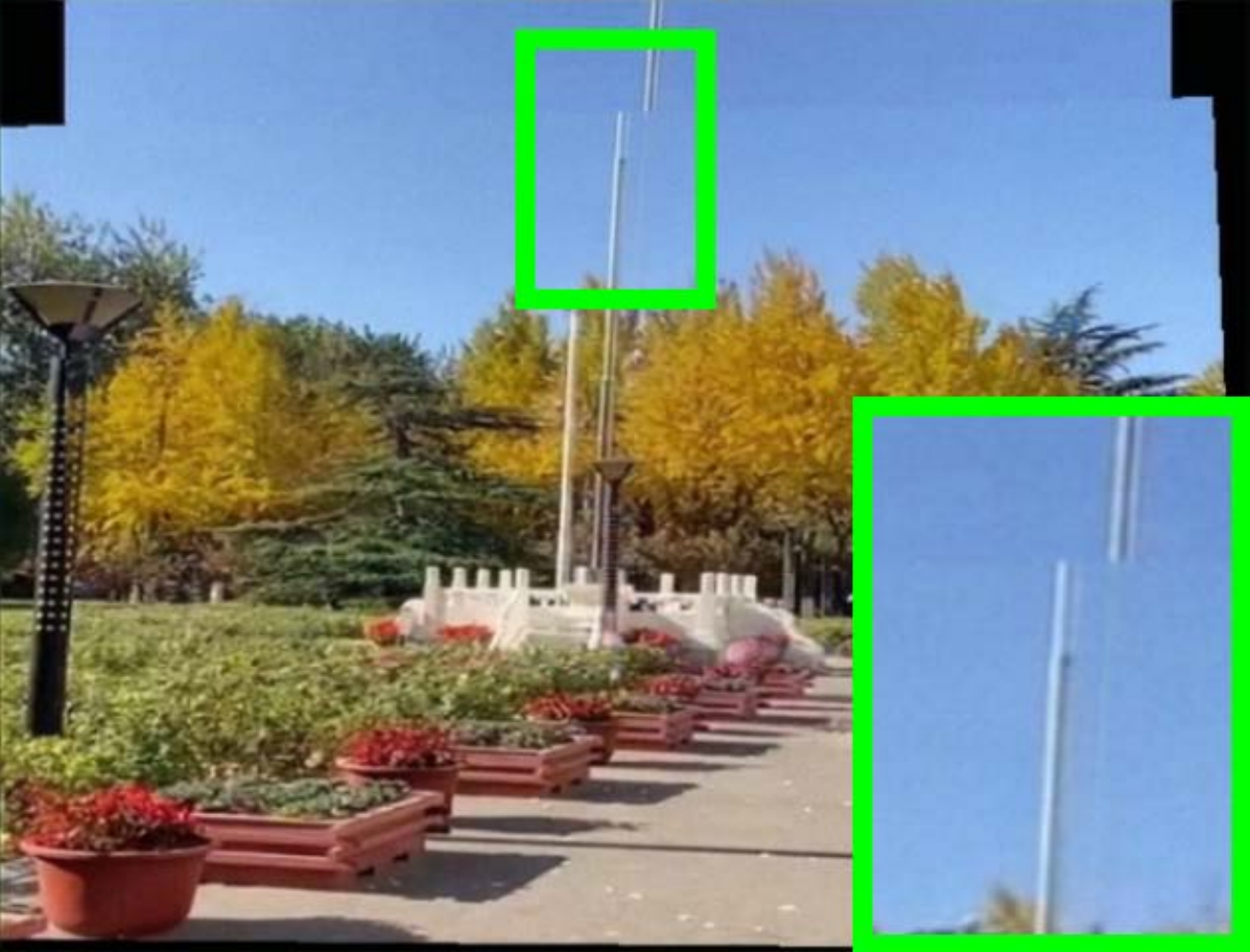}
		&\includegraphics[width=0.13\textwidth,height=0.055\textheight]{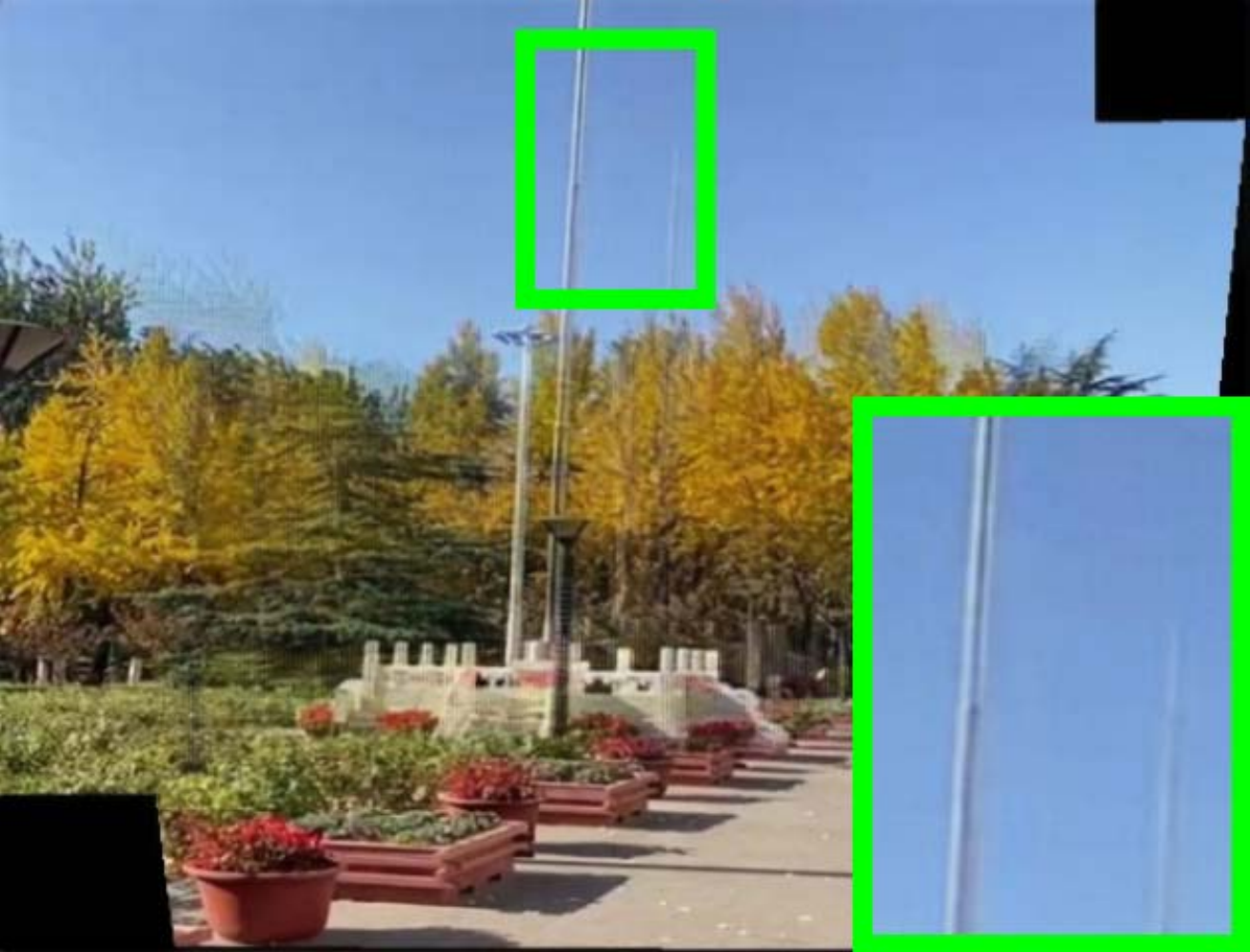}
		&\includegraphics[width=0.13\textwidth,height=0.055\textheight]{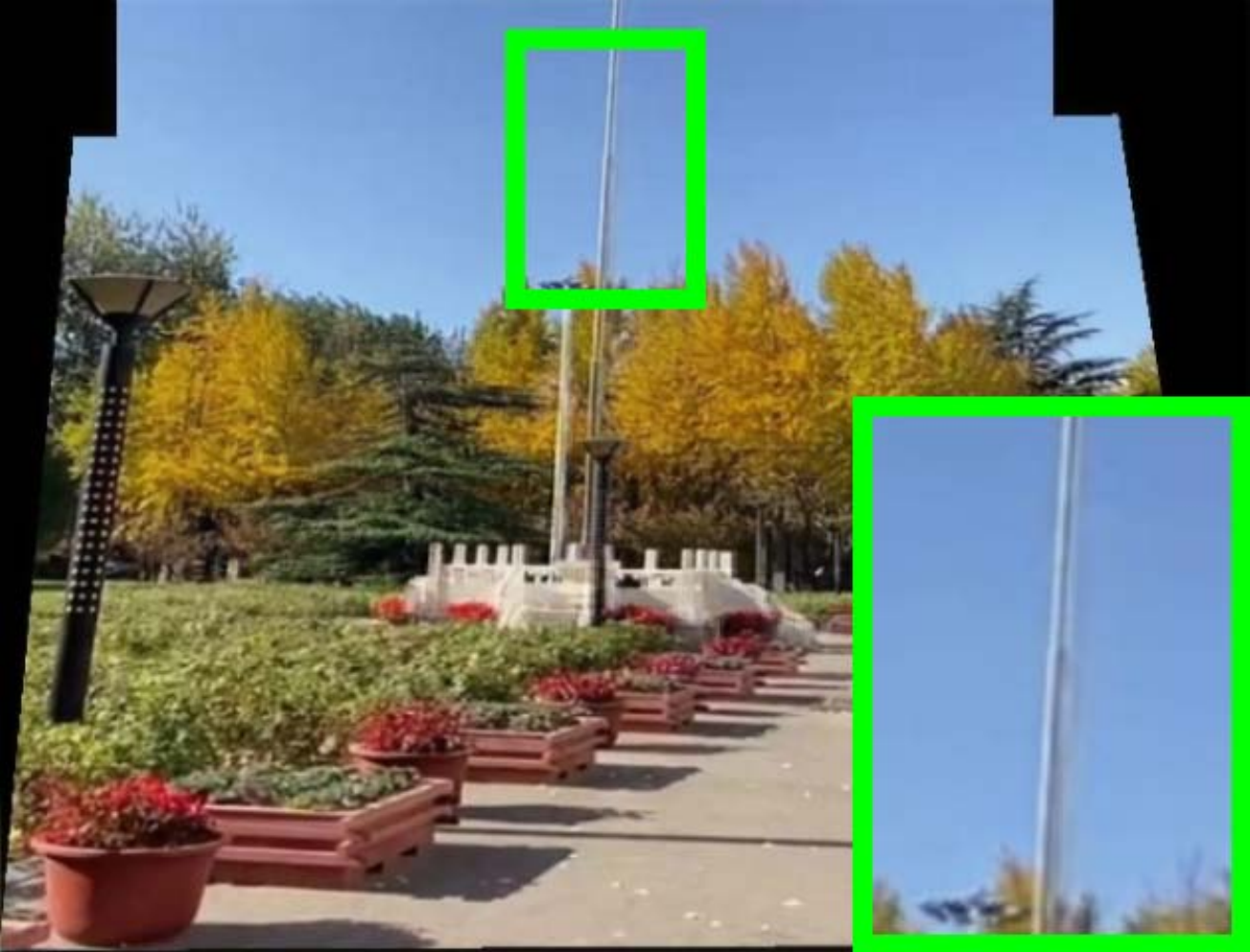}\\
		&Input&Bayesian&Quasi-Newton& Ours\\
	\end{tabular}
	%	\vspace{-.5em}
	\caption{Analysis on the optimization solver of AAT.}
	%	\vspace{-1.5em}
	\label{fig:exp4}
\end{figure}
\begin{figure}[h]
	\centering
	\includegraphics[width=0.45\textwidth]{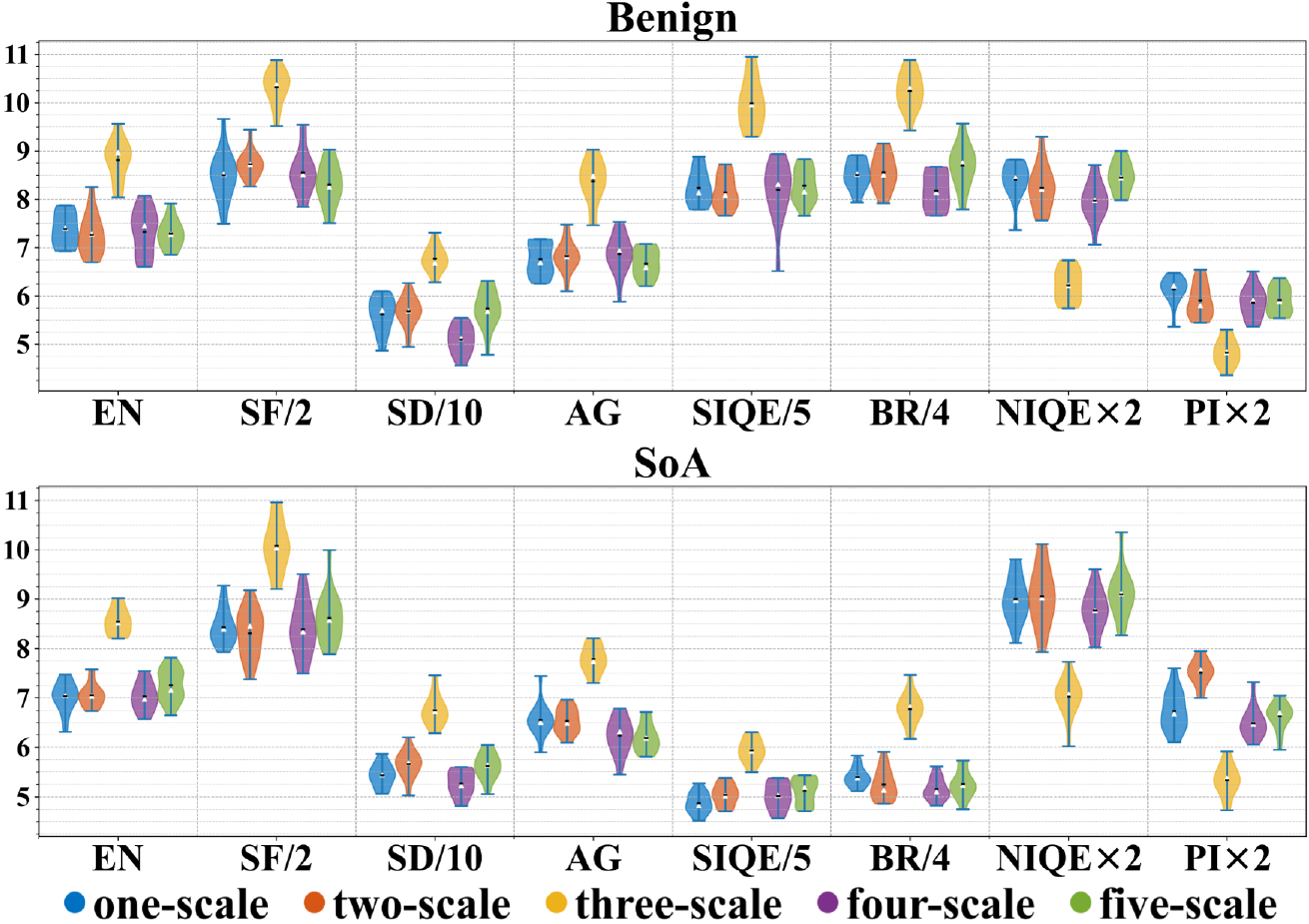}
	%	\vspace{-.5em}
	\caption{Analysis on the robust baseline model in terms of the constrcution of multi-scale features.
%		  Blue, orange, yellow, purple, and green  represent the results of models on a scale of one to five respectively.
	  }
	%	\vspace{-1em}
	\label{fig:exp3}
\end{figure}
\subsubsection{Analysis on the proposed AAT.}
Our AAT employs differentiable architecture search to determine the optimal model, where the updates for architecture parameters are governed by optimization. To achieve optimal performance, we explore various solver strategies within DARTS. We conduct a comparative analysis of model performance derived from different strategies, including Bayesian optimization, quasi-Newton descent, and first-order derivatives~(we used). Visual results are given in Fig.~\ref{fig:exp4}. The model trained via first-order derivatives (denoted as Ours) exhibits superior performance against adversarial attacks. This superiority remains consistent across both benign and adversarial data. The quantitative results presented in Table.~\ref{tab:dataset_comparison} reinforce this observation, with the model derived from the first-order derivatives outperforming others across all evaluation criteria.

\subsubsection{Analysis on the basic multi-scale architecture.}
Benefiting from the hierarchical representation, multi-scale features play a pivotal role in robust perception. Fig.~\ref{fig:exp3} illustrates the comparison of networks with different scales, in which the first figure illustrates the performance on benign data and the second depicts the corresponding performance with SoA attacks. It is evident that the pyramidal structure based on three scales offers not only enhanced stitching performance but also increased resistance to attacks. Consequently, we adopt the three-scale setting for our baseline network.

\begin{figure}[]
	\centering
	\includegraphics[width=0.41\textwidth]{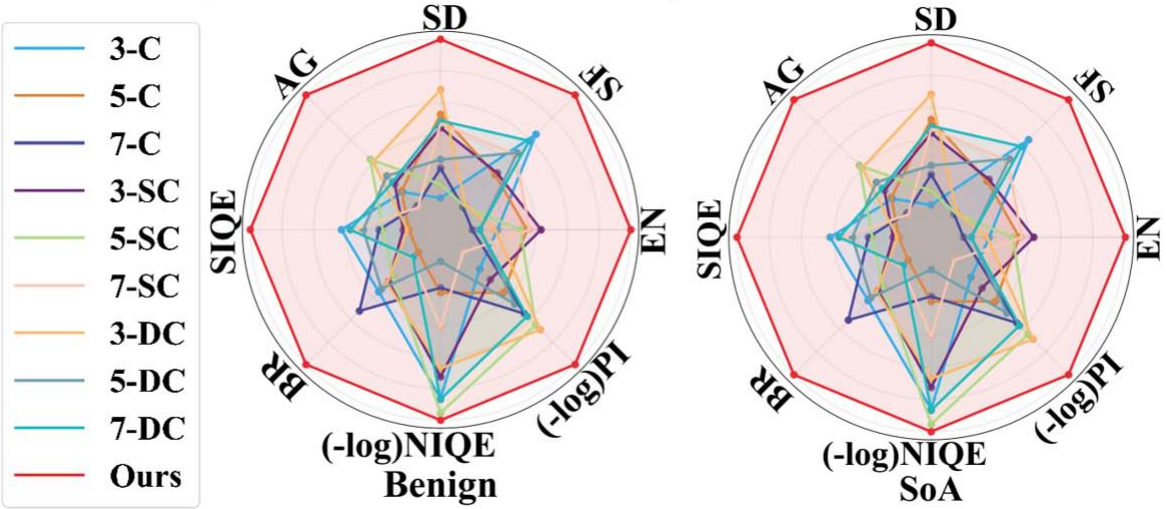}
	%	\vspace{-.5em}
	\caption{Analysis on the candidate operation setting.}
	%	\vspace{-1em}
	\label{fig:leida}
\end{figure}
\subsubsection{Analysis on candidate operation.}
The performance of the final model in neural architecture search is significantly influenced by the choice of candidate operations. We carried out an extensive experimental comparison of candidate operations, including substituting candidate operations within each cell with~$3\times3$ Conv~(3-C),~$5\times5$ Conv~(5-C),~$7\times7$ Conv~(7-C),~$3\times3$ SepConv~(3-SC),~$5\times5$ SepConv~(5-SC),~$7\times7$ SepConv~(7-SC),~$3\times3$ Dilated Conv~(3-DC),~$5\times5$ Dilated Conv~(5-DC), and ~$7\times7$ Dilated Conv~(7-DC). Using our proposed SoA based AAT mechanism, we trained different architectures and identified the one with the best balance of performance and robustness. The evaluation of these models is shown in Fig.~\ref{fig:leida}. Our model demonstrated superior performance on both benign and adversarial data, validating the appropriateness of our proposed candidate operations.

\section{Conclusion}
This paper addressed the vulnerability of deep learning based stitching models against imperceptible perturbations and proposed a robust image stitching method. We introduced a stitching-oriented attack~(SoA) tailored for the alignment of shared regions. Furthermore, we developed an adaptive adversarial training~(AAT) to facilitate the attack-resistant model. During the adversarial training, robust architectures and efficient parameters are automatically determined in an alternative manner. Extensive comparisons with existing learning based image stitching methods and attack strategies demonstrate that the proposed method possesses a strong resilience to perturbations, alleviating the performance disparity between the benign and attacked scenarios. 

\section{Acknowledgments}
This work is partially supported by the National Key R\&D Program of China (Nos. 2020YFB1313500 and 2022YFA1004101), the National Natural Science Foundation of China (Nos. U22B2052 and 62302078), and China Postdoctoral Science Foundation (No. 2023M730741).
%This work is partially supported by the National Key R\&D Program of China under Grants 2020YFB1313500 and 2022YFA1004101, the National Natural Science Foundation of China under Grants U22B2052 and 62302078, and China Postdoctoral Science Foundation under Grant 2023M730741.
\bigskip
%\noindent Thank you for reading these instructions carefully. We look forward to receiving your electronic files!

\bibliography{aaai23}

\end{document}